\documentclass[letterpaper]{article} 
\usepackage{aaai2026}  
\usepackage{times}  
\usepackage{helvet}  
\usepackage{courier}  
\usepackage[hyphens]{url}  
\usepackage{graphicx} 
\usepackage{natbib}  
\usepackage{caption} 
\usepackage{algorithm}
\usepackage{algorithmic}
\usepackage{amsfonts}
\usepackage{enumitem}
\usepackage{amsmath, amsthm, amssymb}
\usepackage{booktabs} 
\usepackage{xcolor}
\usepackage{multirow}
\usepackage{subcaption}
\usepackage{colortbl}

\usepackage{newfloat}
\usepackage{listings}
\DeclareCaptionStyle{ruled}{labelfont=normalfont,labelsep=colon,strut=off} 
\floatstyle{ruled}
\newfloat{listing}{tb}{lst}{}
\floatname{listing}{Listing}
\title{Fair Domain Generalization:  An Information-Theoretic View}
\author {
    Tangzheng Lian\textsuperscript{\rm 1},
    Guanyu Hu\textsuperscript{\rm 2,\rm 3},
    Dimitrios Kollias\textsuperscript{\rm 3},
    Xinyu Yang\textsuperscript{\rm 2},
    Oya Celiktutan\textsuperscript{\rm 1}
}
\affiliations {
    \textsuperscript{\rm 1}King's College London\\
    \textsuperscript{\rm 2}Xi'an Jiaotong University\\
    \textsuperscript{\rm 3}Queen Mary University of London\\
    lian.tangzheng@kcl.ac.uk, g.hu@qmul.ac.uk, d.kollias@qmul.ac.uk, yxyphd@mail.xjtu.edu.cn, oya.celiktutan@kcl.ac.uk 
}

\usepackage{bibentry}

\begin{document}

\maketitle

\begin{abstract}
Domain generalization (DG) and algorithmic fairness are two key challenges in machine learning. However, most DG methods focus solely on minimizing expected risk in the unseen target domain, without considering algorithmic fairness. Conversely, fairness methods typically do not account for domain shifts, so the fairness achieved during training may not generalize to unseen test domains. In this work, we bridge these gaps by studying the problem of Fair Domain Generalization (FairDG), which aims to minimize both expected risk and fairness violations in unseen target domains. We derive novel mutual information-based upper bounds for expected risk and fairness violations in multi-class classification tasks with multi-group sensitive attributes. These bounds provide key insights for algorithm design from an information-theoretic perspective. Guided by these insights, we propose a practical method that solves the FairDG problem through Pareto optimization. Experiments on real-world vision and language datasets show that our method achieves superior utility–fairness trade-offs compared to existing approaches.
\end{abstract}


\section{Introduction}

In real-world deployments, machine learning models often face domain shift, where test data comes from a domain that \textbf{\textit{never seen}} during training (e.g., new demographics, lighting conditions, or image styles). Domain generalization (DG) tackles a more challenging setting, aiming to train models that perform well on unseen domains without access to their data or labels. Instead, it typically assumes the availability of multiple distinct but related source domains during training. Prior DG research has proposed various techniques, including domain-invariant representation learning \cite{ganin2016domain}, data augmentation \cite{dunlap2023diversify}, and meta-learning \cite{li2018learning}. However, these methods focus solely on minimizing expected risk in the target domain and overlook algorithmic fairness. As a result, models that generalize well may still exhibit unfairness in unseen domains.


\begin{figure}[t]
  \centering
  \includegraphics[height=7.2cm, width=8.4cm]{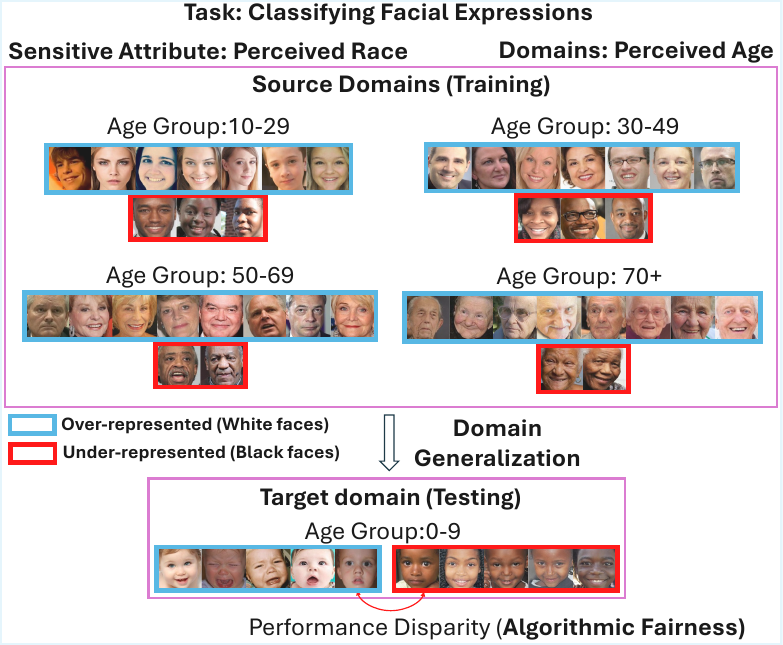}
\caption{A real-world example of the FairDG problem. The goal is to train a model that generalizes to an unseen domain (age group: 0-9) while also ensuring fairness by minimizing performance disparities across perceived racial groups.}
\end{figure}

In parallel, the field of algorithmic fairness in machine learning focuses on mitigating biases in model predictions. Among fairness notions like individual and counterfactual fairness, we focus on the widely used \textbf{\textit{group fairness}} \cite{Caton}, which aims to prevent performance disparities across subgroups defined by a sensitive attribute. These disparities often arise from imbalances in the training data. For instance, as shown in Fig. \textcolor{red}{1}, if white faces dominate the training data while black faces are under-represented, a model trained for facial expression recognition may achieve higher accuracy for white faces and lower accuracy for black faces simply because white faces are more frequent in the training set. Many methods have been proposed to enforce group fairness, typically categorized into pre-processing, in-processing, and post-processing techniques \cite{mehrabi2021survey}. However, these methods generally do not account for domain shifts, so the fairness achieved during training may not generalize to unseen test domains. In this paper, we bridge these gaps by addressing the challenge of Fair Domain Generalization (FairDG), which aims to jointly minimize expected risk and fairness violations in unseen target domains. Our contributions are summarized as follows:

(1) We derive novel theoretical upper bounds based on mutual information (MI) for both the expected risk and fairness violations in multi-class classification tasks with multi-group sensitive attributes, offering key insights from an information-theoretic perspective that inform algorithm design for solving the FairDG problem.

(2) We propose a method to solve FairDG while modeling the utility-fairness trade-off via Pareto optimization.

(3) Experimental results on real-world natural language and vision datasets show that our method outperforms existing approaches, achieving better utility-fairness trade-offs.


\section{Related Works}

The FairDG problem lies within the broader area of ensuring algorithmic fairness under distribution shifts. Please see \cite{shao2024supervised, Barrainkua2025preserving} for comprehensive surveys. However, most prior work focuses on fairness in the domain adaptation, which assumes access to unlabeled target data to adapt to the domain shifts \cite{chen2022fairness, wang2023robust, rezaei2021robust,singh2021fairness}. In contrast, only a few studies have addressed fairness in the more challenging DG setting as discussed below.

Lin et al. proposed two approaches to FairDG: one focusing on group fairness \cite{lin2024fade} and the other on counterfactual fairness \cite{lin2024to}. However, both methods were evaluated only on synthetic and tabular datasets, leaving their effectiveness on real-world, high-dimensional data such as text and images unclear. \cite{jiang2024feed} introduced a meta-learning method, while \cite{tian2024fairdomain} proposed a plug-and-play fair identity attention module for medical image segmentation and classification. \cite{zhao2024algorithmic} addressed FairDG using synthetic data augmentation with learned transformations, and \cite{palakkadavath2025fair} studied the setting with heterogeneous sensitive attributes across domains. Although these methods involve real-world data such as images, none provide theoretical guarantees to support their algorithmic designs. In contrast, our work offers upper bounds on both expected risk and fairness violations and presents a practical framework validated on real-world natural language and vision datasets.

\cite{pham2023fairness} is the first work in FairDG that proposes a method with upper bounds for both fairness violations and expected risk. However, their bounds scale poorly as the number of classes, source domains, and attribute groups increases. In particular, their fairness bound applies only to binary classification with binary-group sensitive attributes and relies on matching distribution means, which is not sufficient to satisfy group fairness metrics. In contrast, we introduce novel bounds based on MI that scale to complex FairDG settings and align directly with the definitions of group fairness metrics. These bounds provide information-theoretic insights that perfectly support algorithm design for FairDG. A detailed comparison with prior theoretical bounds is provided in \textbf{Appendix~\textcolor{red}{D}}.

\section{Problem Formulation}


\begin{table}[ht]
    \centering
    \setlength{\tabcolsep}{2pt}
\renewcommand{\arraystretch}{1.12}
{\fontsize{9pt}{9pt}\selectfont
    \begin{tabular}{ll}
        \hline
        \textbf{Symbol} & \textbf{Description} \\
        \hline
        $\mathcal{X}$          & Input space \\
        $\mathcal{Y}$          & Set of class labels \\
        $\mathcal{D}_S$        & Set of source domains available during training \\
        $\mathcal{G}$          & Set of group memberships (sensitive attribute $S$) \\
        $\mathbf{X} \in \mathcal{X}$ & Random input \\
        $\mathbf{Y} \in \mathcal{Y}$ & Class label corresponding to $\mathbf{X}$ \\
        $\mathbf{D}_S \in \mathcal{D}_S$ & Source domain corresponding to $\mathbf{X}$ \\
        $\mathbf{G} \in \mathcal{G}$ & Group membership corresponding to $\mathbf{X}$ \\
        $d_T$                  & Unknown target domain to generalize \\
        $x$, $y$, $d_S$, $g$  & Realizations of $\mathbf{X}$, $\mathbf{Y}$, $\mathbf{D}_S$, $\mathbf{G}$ \\
        \hline
    \end{tabular}
    }
    \caption{Notation table.}
\end{table}

\noindent\textbf{Assumption 1}. There exists a domain random variable $
\mathbf D \sim\mathrm{Categorical}\bigl(\{\pi_d\}_{d\in\mathcal D}\bigr)$, where $\mathcal{D}$ contains source domains $\mathcal{D}_S$ with $|\mathcal{D}_S| \geq 2$ and an unseen target domain $d_T\notin \mathcal{D}_S$. 

\noindent \textbf{Assumption 2}. We focus on the same sensitive attribute \(S\) and its groups \(\mathcal{G}\) when moving from any \(d_S \in \mathcal{D}_S\) to \(d_T\).  

\noindent\textbf{Domain Generalization:} Let $\hat f_\theta:\mathcal X\to\mathcal Y$ be a model parameterized by $\theta\in\Theta$ and let \( \mathcal{L}(\cdot) \) denote a loss function. The objective of domain generalization is to find the set of  optimal parameters $\theta_{\mathrm{DG}}$ that satisfies:
\begin{equation}
\theta_{\mathrm{DG}}
\;=\;
\arg\min_{\theta\in\Theta}\;\mathcal{R}_{d_T}(\hat{f}_\theta),
\end{equation}
where $\mathcal{R}_{d_T}(\hat{f}_\theta)=\mathbb{E}_{(\mathbf{X},\mathbf{Y})\sim d_T}[\mathcal{L}(\hat{f}_\theta(\mathbf{X}), \mathbf{Y})]$ is the expected risk on an unseen target domain $d_T$.

\noindent\textbf{Algorithmic Fairness:} We consider two group fairness metrics that are conditioned on the true label $\mathbf{Y}$: \textit{Equalized Odds (EOD)} and \textit{Equal Opportunity (EO)}\footnote{Other metrics like demographic parity and disparate impact do not consider the correctness of model predictions. Conditioning on the true label allows fairness evaluation via the confusion matrix, better supporting decision-making \cite{hardt2016equality}.}. In this paper, we focus on deriving EOD, since EO is a special case of EOD that only considers the true positive rate, and \textbf{\textit{all the theoretical results for EOD can be naturally extended to EO.}}

Let $\hat{\mathbf{Y}} = \hat f_\theta(\mathbf{X})$ be the model prediction for a random input, and let $\hat{y} \in \mathcal{Y}$ denote its realization. EOD requires that, for any true label \( y \in \mathcal{Y} \) and any pair of distinct groups \( g ,g' \in \mathcal{G}\), the conditional distributions of the predictions—both true and false positive rates—are identical:
\[
P({\hat{\mathbf{Y}} \mid \mathbf{Y}=y,\, \mathbf{G}=g}) = P({\hat{\mathbf{Y}} \mid \mathbf{Y}=y,\, \mathbf{G}=g'}),
\]
which we denote as $P_{\hat{\mathbf{Y}} \mid y,\, g} = P_{\hat{\mathbf{Y}} \mid y,\, g'}$.  Equivalently, in probability mass functions, this condition is expressed as: $p(\hat{y} \mid y,\,g)=p(\hat{y} \mid y,\,g') \; \forall \hat{y}$. Violations of EOD are measured using the Total‐Variation (TV) distance, defined as: 
$$\delta_{\mathrm{TV}}\!(P_{\hat{\mathbf{Y}} \mid y,\, g}, P_{\hat{\mathbf{Y}} \mid y,\, g'})
=\frac{1}{2}\sum_{\hat{y}\in\mathcal Y}|\,p(\hat{y}\mid y,g)\;-\;p(\hat{y}\mid y,g')|.$$
The overall EOD violation averaging across all classes and group pairs for a model $\hat{f}_\theta$ is then given by:
\[
\Delta^{\text{EOD}} (\hat{f}_\theta) = C^{\text{D}} \sum_{y \in \mathcal{Y}}  \sum_{\{g, g'\} \subset \mathcal{G}} \delta_{\mathrm{TV}}\!(P_{\hat f_\theta(\mathbf{X}) \mid y,\, g}, P_{\hat f_{\theta}(\mathbf{X}) \mid y,\, g'}),
\]

\noindent where the normalization constant $C^{\text{D}} = \frac{2}{|\mathcal{Y}||\mathcal{G}|(|\mathcal{G}| - 1)}$, and the summation is over all unordered group pairs $\{g, g'\}$. The objective is to find the optimal parameters $\theta_{\mathrm{Fair}}$ that minimize the EOD violation on the unseen target domain $d_T$:
\begin{equation}
\theta_{\mathrm{Fair}}
\;=\;
\arg\min_{\theta\in\Theta}\;\Delta^{\text{EOD}}_{d_T} (\hat{f}_\theta).
\end{equation}

\section{Theoretical Bounds}

Minimizing Eq.~(\textcolor{red}{1}) and (\textcolor{red}{2}) is challenging as the target domain $d_T$ is unknown. Therefore, we derive their upper bounds and minimize the bounds instead. (See proofs of the theorems and supporting lemmas from \textbf{Appendix \textcolor{red}{A}} to \textcolor{red}{C}.) 

\noindent\textbf{Theorem 1 (upper bound for the expected risk on the target domain).} Let $\mathcal{L}(\cdot)$ be any non-negative loss function upper bounded\footnote{For example, the CE loss can be bounded by $C$ by modifying the softmax output from $(p_1, p_2, \cdots, p_{|\mathcal{Y}|})$ to $(\hat{p}_1, \hat{p}_2, \cdots, \hat{p}_{|\mathcal{Y}|})$, where $\hat{p}_i = p_i (1 - \exp(-C)|\mathcal{Y}|) + \exp(-C), \ \forall i \in |\mathcal{Y}|$.} by a constant $C$. Then the expected risk on the target domain $d_T$ satisfies the following upper bound:
\[
\begin{aligned}
\mathcal{R}_{d_T}(\hat{f}_\theta) &\leq 
\underbrace{\mathcal{R}_{\mathcal{D}_S}(\hat{f}_\theta)}_{\textbf{Term (1)}} + 
\underbrace{C \cdot \delta_{TV}(P^{\mathbf{X},\mathbf{Y}}_{d_T}, P^{\mathbf{X}, \mathbf{Y}})}_{\textbf{Term (2)}} \\&\hspace{2.5cm}+ 
\underbrace{\frac{\sqrt{2}C}{2}\sqrt{I\Bigl(\mathbf{D}_S;\, \hat{f}_\theta(\mathbf{X}), \mathbf{Y}\Bigr)}}_{\textbf{Term (3)}}
\end{aligned}
\]


\textbf{Term (1)} is the expected risk of the source domains \(\mathcal{D}_S\) available during training. \textbf{Term (2)} is the discrepancy between the joint distribution of inputs and labels in the target domain and the mixture distribution, measured by the TV distance. The mixture distribution is computed from the training data as: \(P^{\mathbf X,\,\mathbf Y}=
\sum_{d_S \in \mathcal{D}_S} p(d_S)P^{\mathbf X,\,\mathbf Y}_{d_S}\). However, in DG, the target domain distribution is unknown, making \textbf{Term (2)} uncontrollable. \textbf{Term (3)} is the MI between the source domain variable \( \mathbf{D}_S \) and the joint variable of the model prediction and label $(\hat{f}_\theta(\mathbf{X}), \mathbf{Y})$, which can then be factorized by the chain rule as: $I(\mathbf{D}_S; \hat f_\theta(\mathbf{X}),\mathbf{Y})
=I(\mathbf{D}_S;\mathbf{Y})+I(\mathbf{D}_S;\hat f_\theta(\mathbf{X})|\mathbf{Y})$, where $I(\mathbf{D}_S;\mathbf{Y})$ is constant and can be estimated from the training data.

\noindent\textbf{Takeaways:} To minimize $\mathcal{R}_{d_T}(\hat{f}_\theta)$, one should focus on minimizing the \textbf{\textit{controllable}} and \textbf{\textit{parameterized}} components of the upper bound: $\mathcal{R}_{\mathcal{D}_S}(\hat{f}_\theta)$ and $I(\mathbf{D}_S;\hat f_\theta(\mathbf{X})|\mathbf{Y})$.

\noindent\textbf{Theorem 2 (upper bound for the EOD violation on the target domain).} The EOD violation for multi-class classification with a multi-group sensitive attribute on the target domain $d_T$ satisfies the following upper bound:
\[
\begin{aligned}
 \Delta_{d_T}^{\mathrm{EOD}}(\hat{f}_\theta) &\leq 
\underbrace{\frac{\sqrt{2I\bigl(\mathbf G;\,\hat f_\theta(\mathbf{X})\mid\mathbf Y,\mathbf D_S\bigr)}}{|\mathcal{Y}||\mathcal{G}|\displaystyle \min_{\substack{y, g}}p(y,g)}}_{\textbf{Term (1)}} \\&+ 
\underbrace{\frac{2}{|\mathcal{Y}||\mathcal{G}|} \sum_{y \in \mathcal{Y}} \sum_{g \in \mathcal{G}} 
\delta_{\mathrm{TV}}\bigl(P^{\mathbf{X} \mid y,g}_{d_T},\, P^{\mathbf{X} \mid y,g}\bigr)}_{\textbf{Term (2)}} \\&\hspace{2.5cm}+ 
\underbrace{\frac{\sqrt{2I\bigl(\mathbf D_S;\,\hat f_\theta(\mathbf{X})\mid\mathbf Y,\mathbf G\bigr)}}{|\mathcal{Y}||\mathcal{G}|\displaystyle \min_{\substack{y, g}}p(y,g)}}_{\textbf{Term (3)}}
\end{aligned}
\]

\textbf{Term (1)} is the MI between the group variable $\mathbf{G}$ and the model prediction $\hat{f}_\theta(\mathbf{X})$, conditioned on the joint variable of the label and source domain $(\mathbf{Y}, \mathbf{D}_S)$. The denominator includes $p(y, g)$, the joint probability of observing label $y$ and group $g$, which is constant and can be estimated from the training data. Similar to \textbf{Theorem 1}, in \textbf{Term (2)}, $P^{\mathbf{X} \mid y,g}$ is computed from the training data by \(P^{\mathbf{X} \mid y,g}=
\sum_{d_S \in \mathcal{D}_S} p(d_S)P^{\mathbf{X} \mid y,g}_{d_S}\). However, the target domain distribution is unknown in DG, making \textbf{Term (2)} uncontrollable. \textbf{Term (3)} measures the MI between the source domain variable $\mathbf{D}_S$ and the model prediction $\hat{f}_\theta(\mathbf{X})$, conditioned on the joint variable of the label and group $(\mathbf{Y}, \mathbf{G})$. 

\noindent\textbf{Takeaways:} To reduce $\Delta_{d_T}^{\mathrm{EOD}}(\hat{f}_\theta)$, one should focus on minimizing the the two \textbf{\textit{parameterized}} conditional MI terms $I\big(\mathbf{G};\, \hat{f}_\theta(\mathbf{X})|\mathbf{Y}, \mathbf{D}_S\big)$ and $I\big(\mathbf{D}_S;\, \hat{f}_\theta(\mathbf{X}) | \mathbf{Y}, \mathbf{G}\big)$.

\section{An Information-Theoretic View}

Combining the two takeaways from \textbf{Theorems 1\&2}, the FairDG objectives can be summarized as finding the optimal parameter set $\theta^*$ that minimizes the following four terms:
\begin{equation}
\begin{aligned}
\hspace{-0.1cm}\theta^*=
&\arg\min\limits_{\theta \in \Theta} \;\{ \mathcal{R}_{\mathcal{D}_S}(\hat{f}_\theta), \;I(\mathbf{D}_S;\hat f_\theta(\mathbf{X})|\mathbf{Y}),\\
& \hspace{0.6cm}I(\mathbf{G};\hat{f}_\theta(\mathbf{X})|\mathbf{Y}, \mathbf{D}_S), I(\mathbf{D}_S; \hat{f}_\theta(\mathbf{X}) | \mathbf{Y}, \mathbf{G})\}.
\end{aligned}
\end{equation}



\noindent\textbf{Theorem 3 (Risk minimization $\Longleftrightarrow$ MI maximization).} When \(\mathcal{L}(\cdot)\) is the cross-entropy loss, the optimal set $\theta^*$ that attain the minimum of $\mathcal R_{\mathcal D_S}(\hat f_\theta)$ (Bayes-optimal) coincides with the set that attains the maximum of $I\!\bigl(\hat f_\theta(\mathbf X);\mathbf Y | \mathbf D_S\bigr)$ (Please refer to detailed proof in the \textbf{Appendix \textcolor{red}{B}}):
$$
\arg\min_{\theta\in\Theta}\;\mathcal R_{\mathcal D_S}(\hat f_\theta)
\;=\;
\arg\max_{\theta\in\Theta}\;
I\!\bigl(\hat f_\theta(\mathbf X)\,;\,\mathbf Y \mid \mathbf D_S\bigr).
$$

Therefore, Eq.~(\textcolor{red}{3}) can be interpreted entirely in MI terms:
\begin{equation}
\begin{aligned}
\hspace{-0.1cm}\theta^*=
&\arg\max\limits_{\theta \in \Theta} I(\hat{f}_\theta(\mathbf{X});\, \mathbf{Y} \mid \mathbf{D}_S), \\
&\arg\min\limits_{\theta \in \Theta} \{I(\mathbf{D}_S;\hat f_\theta(\mathbf{X})|\mathbf{Y}),I(\mathbf{G};\hat{f}_\theta(\mathbf{X})|\mathbf{Y}, \mathbf{D}_S), \\
& \hspace{3.5cm}I(\mathbf{D}_S;\, \hat{f}_\theta(\mathbf{X}) \mid \mathbf{Y}, \mathbf{G})\}.
\end{aligned}
\end{equation}



\noindent\textbf{Theorem 4 (Chain-rule bounds).} For the random variables $\mathbf{X}$, $\mathbf{Y}$, $\mathbf{D}_S$, $\mathbf{G}$, and for any parameter set $\theta$, the MI terms in Eq. (\textcolor{red}{4}) satisfy the following inequalities based on chain rules (Please refer to detailed proof in the \textbf{Appendix \textcolor{red}{B}}):
\begin{equation}
    \begin{aligned}
I(\mathbf{D}_S;\hat{f}_\theta(\mathbf{X})| \mathbf{Y},\mathbf{G})
  &\leq I(\mathbf{D}_S;\hat{f}_\theta(\mathbf{X})| \mathbf{Y})
  \\&\hspace{1.2cm}+ I(\mathbf{G};\, \hat{f}_\theta(\mathbf{X}) | \mathbf{Y}, \mathbf{D}_S), 
\end{aligned}
\end{equation}
\begin{equation}
\hspace{-0.1cm}I\bigl(\hat f_\theta(\mathbf X);\mathbf Y\bigr)\geq 
I\bigl(\hat f_\theta(\mathbf X);\mathbf Y| \mathbf D_S\bigr)-
I\bigl(\mathbf D_S;\hat f_\theta(\mathbf X)| \mathbf Y\bigr),
\end{equation}
\begin{equation}
    \begin{aligned}
I(\mathbf G;\hat f_\theta(\mathbf X)| \mathbf Y)&\leq I(\mathbf D_S;\hat f_\theta(\mathbf X)| \mathbf Y)
\\&\hspace{1.9cm}+
I(\mathbf G;\hat f_\theta(\mathbf X)| \mathbf Y,\mathbf D_S). 
\end{aligned}
\end{equation}

Eq.~(\textcolor{red}{5}) shows that for any parameter set \(\theta\), the last MI term in Eq.~(\textcolor{red}{4}) is upper-bounded by the sum of the second and third MI terms. Hence, reducing the second and third MI terms tightens this upper bound and can already help to decrease the last MI term. We can then simplify Eq.~(\textcolor{red}{4}) to:
\begin{equation}
\begin{aligned}
\hspace{-0.14cm}\theta^*=
&\arg\max\limits_{\theta \in \Theta} I(\hat{f}_\theta(\mathbf{X});\, \mathbf{Y} \mid \mathbf{D}_S), \\
&\arg\min\limits_{\theta \in \Theta} \{I(\mathbf{D}_S;\hat f_\theta(\mathbf{X})|\mathbf{Y}),I(\mathbf{G};\hat{f}_\theta(\mathbf{X})|\mathbf{Y}, \mathbf{D}_S)\}
\end{aligned}
\end{equation}

We further show that Eq.~(\textcolor{red}{8}) indeed provides a solution to the FairDG problem from an information-theoretic view. For DG,  Eq.~(\textcolor{red}{6}) indicates that maximizing $I(\hat f_\theta(\mathbf X);\mathbf Y| \mathbf D_S)$ and minimizing $I(\mathbf D_S;\hat f_\theta(\mathbf X)| \mathbf Y)$ in the Eq.~(\textcolor{red}{8}) raises the lower bound of $I(\hat f_\theta(\mathbf X);\mathbf Y)$, thereby helping to increase it. This aligns with the goal of DG: by increasing $I(\hat f_\theta(\mathbf X);\mathbf Y)$, the model learns to make predictions more informative about the true labels regardless of source domains. As a result, the model becomes source domain-invariant and is better positioned to generalize to unseen target domains. Similarly, for algorithmic fairness, Eq.~(\textcolor{red}{7}) indicates that minimizing both $I(\mathbf D_S;\hat f_\theta(\mathbf X)| \mathbf Y)$ and $I(\mathbf G;\hat f_\theta(\mathbf X)| \mathbf Y,\mathbf D_S)$ in the Eq.~(\textcolor{red}{8}) reduces the upper bound of $I(\mathbf G;\hat f_\theta(\mathbf X)| \mathbf Y)$, thereby help to decrease it. This aligns with our goal of algorithmic fairness based on EOD: $I(\mathbf G;\hat f_\theta(\mathbf X)| \mathbf Y)$ characterizes the MI form of EOD violations regardless of source domains. Minimizing this term encourages the model to produce EOD-consistent predictions that are source domain-invariant, thereby enhancing its ability to generalize the EOD-based algorithmic fairness to unseen target domains.

\section{Proposed Method}
Although Eq. (\textcolor{red}{8}) offers a theoretical formulation of the FairDG problem from an information-theoretic perspective, the direct computation of the MI terms is impractical as the underlying probability distributions of the involved random variables are unknown. To address this, as shown in Fig. \textcolor{red}{2}, we introduce a practical method designed to approximate and optimize Eq. (\textcolor{red}{8}) with finite training data. 

First, directly optimizing the predicted label $\hat{f}_\theta(\mathbf{X})$ is infeasible due to non-differentiable discrete operations (e.g., argmax over logits). A common strategy in fair or domain-invariant representation learning is decomposing the $\hat{f}_\theta$ into a feature encoder \(\hat{f}_{\theta_E}\) and a classifier \(\hat{f}_{\theta_C}\) to enable optimization at the representation level, such that by the data processing inequality the classifier applied after can be fair or domain-invariant \cite{ganin2016domain,quadrianto2019discovering,dehdashtian2024utility}. As shown in Fig.~\textcolor{red}{2}, \(\hat{f}_{\theta_E}\) maps inputs to representations \(\mathbf{Z}_E = \hat{f}_{\theta_E}(\mathbf{X})\), and the objectives of minimizing \(I(\mathbf{D}_S; \hat{f}_\theta(\mathbf{X}) | \mathbf{Y})\) and \(I(\mathbf{G}; \hat{f}_\theta(\mathbf{X}) | \mathbf{Y}, \mathbf{D}_S)\) can be reformulated as minimizing \(I(\mathbf{D}_S; \mathbf{Z}_E | \mathbf{Y})\) and \(I(\mathbf{G}; \mathbf{Z}_E | \mathbf{Y}, \mathbf{D}_S)\).

However, calculating MI for high-dimensional representations is difficult and often requires approximations like the Mutual Information Neural Estimator (MINE) \cite{pmlr-v80-belghazi18a} or bounds-based methods \cite{poole2019variational}. These methods can introduce approximation errors and may rely on unrealistic assumptions (e.g., assuming $\mathbf{Z}_E$ follows a Gaussian distribution). A more practical alternative is to use a differentiable dependence metric that captures both linear and non-linear dependencies, and works well with high-dimensional random vectors. This dependence metric is then used to enforce conditional independence relations $\mathbf{D}_S\perp\mathbf{Z}_E|\mathbf{Y}$ and $\mathbf{G}\perp \mathbf{Z}_E| \mathbf{Y}, \mathbf{D}_S$. Common choices include the Hilbert-Schmidt Independence Criterion (HSIC) \cite{quadrianto2019discovering, bahng2020learning} and Distance Correlation (dCor) \cite{liu2022fair, zhen2022versatile}. Our experimental results show that dCor consistently outperforms both MINE and HSIC, making it the preferred choice for implementing our method (see Section \textbf{Experiments} for detailed discussion). Therefore, the optimization goal becomes minimizing two conditional dCor terms: $\operatorname{dCor}(\mathbf{D}_S,\mathbf{Z}_E|\mathbf{Y})$ and $\operatorname{dCor}(\mathbf{G}, \mathbf{Z}_E| \mathbf{Y}, \mathbf{D}_S)$.

In real-world settings, \(\mathbf{D}_S\) and \(\mathbf{G}\) are usually discrete, while \(\mathbf{Z}_E\) is continuous. Unlike previous studies that compute dCor directly between discrete and continuous variables \cite{guo2022learning, zhang2019estimating}. We argue that it is more effective to represent $\mathbf{D}_S$ and $\mathbf{G}$ as continuous vectors because categorical labels fail to capture intra-group similarities in the way latent representations do \cite{zhen2022versatile, bahng2020learning}. Accordingly, as shown in Fig. \textcolor{red}{2}, we introduce two additional encoders: a domain encoder $\hat{f}_{\theta_D}$ and a group encoder $\hat{f}_{\theta_G}$ to extract domain and group representations $\mathbf{Z}_D=\hat{f}_{\theta_D}(\mathbf{X})$ and $\mathbf{Z}_G=\hat{f}_{\theta_G}(\mathbf{X})$ (we have empirically validated this design in the \textbf{Experiments} section). Therefore, our objectives become minimizing $\operatorname{dCor}(\mathbf{Z}_D,\mathbf{Z}_E|\mathbf{Y})$ and $\operatorname{dCor}(\mathbf{Z}_G, \mathbf{Z}_E| \mathbf{Y}, \mathbf{D}_S)$. Given a training set with $n$ samples $\mathcal{D}_{train}=\{(x_i, y_i, d^{\,i}_S, g_i)\}_{i=1}^n$, the empirical version of these objectives are:
\begin{equation}
\operatorname{dCor}_n(\hat f_{\theta_D}(x_i), \hat f_{\theta_E}(x_i)|y)
\end{equation}
and 
\begin{equation}
\operatorname{dCor}_n(\hat f_{\theta_G}(x_i), \hat f_{\theta_E}(x_i)|y,d_S).    
\end{equation}
\begin{figure}[t]
  \centering
\includegraphics[height=8.4cm, width=6.8cm]{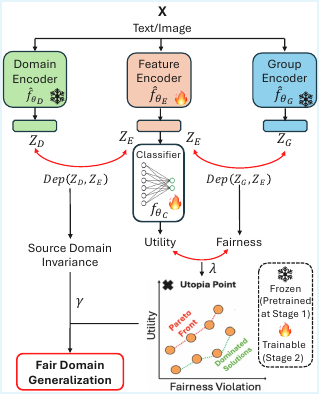}
  \caption{Our proposed method.  \(\hat{f}_{\theta_D}\) and  \(\hat{f}_{\theta_G}\) are trained in stage 1 and frozen to produce $Z_D$ and $Z_G$, which guide  \(\hat{f}_{\theta_E}\) to learn fair and domain-invariant $Z_D$ by minimizing two dependence terms, where $\lambda$ controls the utility–fairness trade-off and $\gamma$ adjusts the source domain invariance.}
\end{figure}
Here, $\operatorname{dCor}_n$ ranges from 0 to 1, with \(\operatorname{dCor}_{n} = 0\) indicating no observable dependence among the samples. Similar to the empirical risk minimization (ERM) in Eq. (\textcolor{red}{11}), $\operatorname{dCor}_{n}$ almost surely converges to the population value as \( n \to \infty \) (see Theorem 2 in \cite{szekely2007measuring}). Full derivations of Eq. (\textcolor{red}{9}) and Eq. (\textcolor{red}{10}) are provided in the \textbf{Appendix \textcolor{red}{E}}. In parallel, as implied by \textbf{Theorem 3}, maximizing $I(\hat{f}_\theta(\mathbf{X});\mathbf{Y}| \mathbf{D}_S)$ can be achieved by minimizing the expected risk over the source domains $\mathcal{R}_{\mathcal{D}_S}(\hat{f}_\theta)$, which reduces to ERM under the training data $\mathcal{D}_{train}$:
\begin{equation}
\frac{1}{n}\sum_{i=1}^n \mathcal{L}(\hat f_{\theta_C}(\hat f_{\theta_E}(x_i)), y_i)
\end{equation}
As prior fairness research shows \cite{taufiq2024achievable, sadeghi2022tradeoff, dehdashtian2024utility}, there is often a trade-off between utility and fairness. Thus, the objectives in Eq. (\textcolor{red}{10}) and Eq. (\textcolor{red}{11}) may conflict. This means no single optimal set $\theta^*$ can simultaneously satisfy $\theta_{\mathrm{DG}}$ and $\theta_{\mathrm{Fair}}$. Instead, the problem should be framed as a multi-objective optimization (MOO) and optimized to yield Pareto-optimal solutions. By combining Eq. (\textcolor{red}{9}), Eq. (\textcolor{red}{10}), and Eq. (\textcolor{red}{11}) with linear scalarization\footnote{Linear scalarization ensures Pareto-optimal solutions, but as both the encoders and the classifier are non-convex, it may not fully characterize the Pareto front \cite{martinez2020minimax}. }, we formulate the empirical objective as:
\begin{equation}
\begin{aligned}
\theta^{\mathcal{P}^*} = \mathop{\arg \min}\limits_{\substack{\theta_E \in \Theta_E \\ \theta_C \in \Theta_C \\ \theta_D \in \Theta_D \\\theta_G \in \Theta_G}} &\; \underbrace{\frac{1-\lambda}{n}\sum_{i=1}^n \mathcal{L}(\hat f_{\theta_C}(\hat f_{\theta_E}(x_i)), y_i)}_{\textbf{Utility (ERM)}} \\&\hspace{-0.2cm}+ \lambda \underbrace{\operatorname{dCor}_n(\hat f_{\theta_G}(x_i), \hat f_{\theta_E}(x_i)|y,d_S)}_{\textbf{Fairness (EOD)}} \\&+ \gamma \underbrace{\operatorname{dCor}_n(\hat f_{\theta_D}(x_i), \hat f_{\theta_E}(x_i)|y)}_{\textbf{Source Domain Invariance}}.
\end{aligned}
\end{equation}

Here, $\lambda \in [0,1)$ balances the utility-fairness trade-off and $\gamma$ controls the strength of the regularization for source domain invariance. We set the upper bound \(C = 1\) for the loss function \(\mathcal{L}(\cdot)\) (CE loss) as described in footnote \textcolor{red}{2}.


\section{Training \& Evaluation}
\textbf{Training:}  A key challenge in optimizing Eq.~(\textcolor{red}{11}) is training stability, as the framework includes four network components. To address this, as shown in Fig. \textcolor{red}{2}, we adopt a two-stage training procedure. Since we have both domain and group labels in the training set, in the first stage, we train $\hat{f}_{\theta_D}$ and $\hat{f}_{\theta_G}$ by attaching classification heads to predict the source domains and group memberships of training samples.  As $\hat{f}_{\theta_D}$ and $\hat{f}_{\theta_G}$ are only used to train $\hat{f}_{\theta_E}$ in a way that it learns to encode $\mathbf{Z}_{E}$ to be conditionally independent of $\mathbf{Z}_{D}$ and $\mathbf{Z}_{G}$, $\hat{f}_{\theta_D}$ and $\hat{f}_{\theta_G}$ are \textbf{\textit{discarded at inference time.}} Therefore, obtaining $\hat{f}_{\theta_D}$ and $\hat{f}_{\theta_G}$ are simple in this case as we just need to train them to \textbf{\textit{overfit}} the training set so that $\mathbf{Z}_{D}$ and $\mathbf{Z}_{G}$ are nearly the optimal representation of $\mathbf{D}_S$ and $\mathbf{G}$ for training samples. In the second stage, we freeze $\hat{f}_{\theta_D}$ and $\hat{f}_{\theta_G}$ and then train $\hat{f}_{\theta_E}$ and $\hat{f}_{\theta_C}$ for the main task. 

Another challenge stems from the MOO setting: achieving different utility-fairness trade-offs requires training a new model from scratch for each $\lambda$, which is computationally expensive. To address this, we adopt the loss-conditional training strategy proposed in \cite{dosovitskiy2019you}. Instead of training separate models for each $\lambda$, we train a single model that conditions on $\lambda$ during training. Specifically, we sample a range of $\lambda$ values and train the model on $(\mathbf{X}, \lambda)$ input pairs. This enables the network to adapt its behavior depending on the desired trade-off. So during inference, we just pass different $\lambda$ to obtain a model tuned for that particular balance between utility and fairness.

\begin{table*}[ht]
\centering
\caption{The classification tasks, domain variables, domain splits, sensitive attributes, and group memberships for each dataset.}
\renewcommand{\arraystretch}{1}
\setlength{\tabcolsep}{1pt} %
{\fontsize{7.1pt}{10pt}\selectfont
\begin{tabular}{c|c|c|c}
\hline
\textbf{Datasets} & \textbf{CelebA} & \textbf{AffectNet} & \textbf{Jigsaw} \\
\hline
\textbf{Classes} 
& \shortstack{\textbf{Hair Colors:}\\\{\textit{black hair}, \textit{brown hair}, \textit{blond hair}\}} 
& \shortstack{\textbf{Facial Expressions:}\\\{\textit{Happiness}, \textit{Sadness}, \textit{Neutral}, \textit{Fear}, \textit{Anger}, \textit{Surprise}, \textit{Disgust}\}} 
& \shortstack{\textbf{Toxic Levels:}\\\{\textit{non-toxic}, \textit{toxic}, \textit{severe toxic}\}} \\
\hline
\textbf{Domains} 
& \shortstack{\textbf{Hairstyles:}\\\{\textit{wavy hair}, \textit{straight hair}, \textit{bangs}, \textit{receding hairlines}\}} 
& \shortstack{\textbf{Perceived Age Groups:}\\\{0–9, 10–29, 30–49, 50–69, 70+\}} 
& \shortstack{\textbf{Toxicity Types:}\\\{\textit{Obscene}, \textit{Identity attack}, \textit{Insult}, \textit{Threat}\}} \\
\hline
\textbf{Splits} 
& \shortstack{\textit{bangs} = test, \textit{receding hairlines} = val, others = train} 
& \shortstack{0–9 = test, 10–29 = val, others = train} 
& \shortstack{\textit{Identity attack} = test, \textit{Threat} = val, others = train} \\
\hline
\textbf{Groups} 
& \shortstack{\textbf{Intersections of Perceived Gender and Age:}\\\{\textit{male-young}, \textit{female-young}, \textit{male-old}, \textit{female-old}\}} 
& \shortstack{\textbf{Perceived race:}\\\{\textit{White}, \textit{Black}, \textit{East Asian}, \textit{Indian}\}} 
& \shortstack{\textbf{Gender-related terms:}\\\{\textit{male}, \textit{female}, \textit{transgender}\}} \\
\hline
\end{tabular}
}
\label{tab:dataset_summary_compact}
\end{table*}

\begin{table}[t]
\centering
\caption{Comparison of existing methods for \((V_{\text{opt}}, U_{\text{opt}})\) evaluation. DG methods are marked with \textcolor{red}{$\dagger$}, fairness-only methods with \textcolor{red}{$*$}, and FairDG methods with \textcolor{red}{$\dagger*$}. Higher Acc (\%) reflects better utility, while lower EOD (\%) and EO (\%) violations indicate better fairness. The best-performing method is shown in bold and underlined; the second-best is underlined. We report only the mean values here; please refer to the \textbf{Appendix \textcolor{red}{G}} for the full tables including variances.}
\setlength{\tabcolsep}{0.4pt}
\renewcommand{\arraystretch}{1}
{\fontsize{7.3pt}{9pt}\selectfont
\begin{tabular}{c|ccc|ccc|ccc}
\hline
  Dataset &
  \multicolumn{3}{c|}{CelebA} &
  \multicolumn{3}{c|}{AffectNet} &
  \multicolumn{3}{c}{Jigsaw} \\ \hline
  Methods &
  \multicolumn{1}{c|}{Acc \textcolor{red}{$\uparrow$}} &
  \multicolumn{1}{c|}{EOD \textcolor{blue}{\textdownarrow}} &
  \multicolumn{1}{c|}{EO \textcolor{blue}{\textdownarrow}} &
  \multicolumn{1}{c|}{Acc \textcolor{red}{$\uparrow$}} &
  \multicolumn{1}{c|}{EOD \textcolor{blue}{\textdownarrow}} &
  \multicolumn{1}{c|}{EO \textcolor{blue}{\textdownarrow}} &
  \multicolumn{1}{c|}{Acc \textcolor{red}{$\uparrow$}} &
  \multicolumn{1}{c|}{EOD \textcolor{blue}{\textdownarrow}} &
  \multicolumn{1}{c}{EO \textcolor{blue}{\textdownarrow}} \\ \hline
\rowcolor{gray!20} 
ERM (Ours) & 82.8 & 14.2 & 10.5 & 62.8 & 15.0 & 10.1 & 82.4 & 17.8 & 11.1 \\
\rowcolor{gray!20} 
\textcolor{red}{$\dagger$} ERM+SDI (Ours) & \textbf{\underline{89.6}} & 13.5 & 11.5 & \textbf{\underline{69.8}} & 15.2 & 11.1 & \textbf{\underline{87.4}} & 18.8 & 11.6 \\
\textcolor{red}{$\dagger$} DANN & 88.6 & 15.5 & 14.4 & 66.8 & 16.5 & 12.4 & 84.2 & 17.7 & 12.1 \\
\textcolor{red}{$\dagger$} CORAL & 87.6 & 14.6 & 14.5 & 67.3 & 15.6 & 14.5 & 83.1 & 18.6 & 13.2 \\
\textcolor{red}{$\dagger$} MMD-AAE & 88.4 & 16.5 & 13.2 & 68.4 & 17.5 & 15.2 & 84.4 & 16.8 & 10.7 \\
\textcolor{red}{$\dagger$} DDG & 86.9 & 17.2 & 14.2 & \underline{69.4} & 18.2 & 14.2 & 84.7 & 18.5 & 11.9 \\ \hline
\rowcolor{gray!20} 
\textcolor{red}{$*$} ERM+Fair (Ours) & 79.4 & 13.8 & 10.4 & 59.4 & 14.8 & 8.4 & 81.4 & 15.5 & 10.9 \\
\textcolor{red}{$*$} LNL & 72.4 & 13.2 & 10.4 & 61.4 & 13.6 & 9.6 & 77.4 & 17.5 & 7.1 \\
\textcolor{red}{$*$} MaxEnt-ARL & 71.5 & 13.6 & 9.0 & 61.5 & 13.8 & 10.0 & 78.7 & 16.5 & 7.7 \\
\textcolor{red}{$*$} FairHSIC & 82.4 & 12.5 & 9.3 & 62.4 & 12.6 & 9.8 & 79.4 & 15.6 & 9.8 \\
\textcolor{red}{$*$} U-FaTE & 69.5 & 14.1 & 8.2 & 59.5 & 14.6 & 6.4 & 79.7 & 16.1 & 8.7 \\ \hline
\textcolor{red}{$\dagger*$} FEDORA & 85.7 & 10.6 & 8.1 & 65.8 & 10.2 & \underline{5.9} & 84.4 & 12.6 & 6.1 \\
\textcolor{red}{$\dagger*$} FATDM-StarGAN & 85.4 & 10.9 & 6.7 & 65.6 & \underline{9.8} & 6.3 & 85.7 & \underline{12.3} & 5.7 \\
\rowcolor{gray!20} 
\textcolor{red}{$\dagger*$} Ours-S & 84.3 & 11.8 & \underline{5.8} & 64.5 & 10.8 & 6.3 & 84.6 & 12.4 & \underline{4.9} \\
\rowcolor{gray!20} 
\textbf{\textcolor{red}{$\dagger*$} Ours} & \underline{88.7} & \textbf{\underline{8.2}} & \textbf{\underline{3.1}} & 68.9 & \textbf{\underline{8.4}} & \textbf{\underline{4.3}} & \underline{86.3} & \textbf{\underline{9.7}} & \textbf{\underline{3.7}} \\ \hline
\end{tabular}
}
\end{table}

\noindent\textbf{Evaluation:} We evaluate models with different \(\lambda\) values during testing using fairness metric \(V\) (either EO or EOD violations) and utility metric \(U\) (accuracy). Let the solution set be \( \mathcal{S}_{(V,U)} = \{(V_i, U_i) \mid i = 1, 2, \dots, N\} \), where each solution corresponds to a model with a different \(\lambda\). We define the set of solutions that dominate a solution \((V_i, U_i)\) as:
\begin{equation*}
\mathcal{D}_{(i)} =  \left\{ (V_j, U_j) \in \mathcal{S}_{(V,U)} \;\middle|\; ( V_j \leq V_i ) \wedge ( U_j \geq U_i ) \right\}. 
\end{equation*} 

The Pareto front \(\mathcal{P}\), containing all non-dominated (Pareto optimal) solutions \((V_i, U_i) \in \mathcal{S}_{(V,U)}\), is defined as:
\begin{equation*}
\mathcal{P} = \left\{ (V_i, U_i) \in \mathcal{S}_{(V,U)} \;\middle|\; \mathcal{D}_{(i)} = \emptyset \right\}.
\end{equation*}

\textbf{\textit{We evaluate both the full Pareto front and a selected single solution.}} We use the Hypervolume Indicator (HVI) \cite{zitzler2007hypervolume} as the evaluation metric to measure both the convergence and diversity of the Pareto front $\mathcal{P}$. HVI measures the area in the solution space dominated by \(\mathcal{P}\), bounded by a reference point \( R = (V_{\text{ref}}, U_{\text{ref}}) \), where \(V_{\text{ref}} > V_{\max}\) and \(U_{\text{ref}} < U_{\min}\). As the utility and fairness may vary in scale (\(\Delta V = V_{\max} - V_{\min}\), \(\Delta U = U_{\max} - U_{\min}\)), we follow the standard practice \cite{miettinen1999nonlinear, branke2008multiobjective} and normalize both metrics to \([0,1]\) :
\begin{equation*}
\mathcal{P}_{\text{norm}} = \left\{ \left(\frac{V_i - V_{\min}}{\Delta V}, \frac{U_i - U_{\min}}{\Delta U} \right) \;\middle|\; (V_i, U_i) \in \mathcal{P} \right\}.
\end{equation*}

In \(\mathcal{P}_{\text{norm}}\), no solution can have both lower \(V\) and higher \(U\). Thus, the solutions can be sorted as \(\mathcal{P}_{\text{norm}} = \{ (V_1, U_1), \ldots, (V_n, U_n) \}\) with \(V_1 < V_2 < \cdots < V_n\) and \(U_1 < U_2 < \cdots < U_n\). The HVI is calculated as the non-overlapping rectangular area under \(\mathcal{P}_{\text{norm}}\) bounded by \(R\):
\begin{equation}
\text{HVI}(\mathcal{P}_{\text{norm}}) = \sum_{i=2}^{n+1} (V_{i} - V_{i-1}) \times (U_{i-1} - U_{\text{ref}}),
\end{equation}
\noindent where $V_{n+1}=V_{\text{ref}}$ with higher HVI for a better Pareto front.

For the single solution, we argue that preferences should be set by the human decision makers; thus, we do not assume a preference between objectives by default. A well-known approach for selecting the optimal solution when preference information is not provided is the global criterion method \cite{zeleny2012multiple,hwang2012multiple}. Let the utopia point \( U = (V^*, U^*) \) denote the ideal but typically unreachable solution. In the normalized space \(\mathcal{P}_{\text{norm}}\), the optimal solution is the one closest to the utopia point in \(L_2\) distance:
\begin{equation}
\hspace{-0.3cm}(V_{\text{opt}}, U_{\text{opt}}) = \!\!\! \underset{(V_i, U_i) \in \mathcal{P}_{\text{norm}}}{\arg\min}\!\!\! \sqrt{ (V_i - V^*)^2 + (U^* - U_i)^2 }.
\end{equation}

\section{Experiments}

\textbf{Datasets:} Prior works on the FairDG problem use datasets that are either synthetic or tabular \cite{lin2024to, lin2024fade}, or restricted to binary classification tasks \cite{pham2023fairness, tian2024fairdomain}. In contrast, real-world FairDG problems are far more challenging, often involving high-dimensional data (e.g., text or images), multi-class classification, and multi-group sensitive attributes. As shown in Table \textcolor{red}{2}, we use three datasets that reflect these complexities: \textbf{CelebA} \cite{liu2015faceattributes}, \textbf{AffectNet} \cite{mollahosseini2017affectnet}, and \textbf{Jigsaw} \cite{jigsaw-multilingual-toxic-comment-classification}. See details on the datasets in the \textbf{Appendix \textcolor{red}{F}}. 

\noindent\textbf{Implementation Details:}  For CelebA, we used ResNet18 \cite{he2016deep} for all encoders; for AffectNet, we used the Swin Transformer (Base) \cite{liu2021swin}; and for Jigsaw, we employed Sentence-BERT \cite{reimers-2019-sentence-bert} for all encoders. All classifiers were implemented as two-layer MLPs.  Training was performed using SGD, and the trade-off parameter $\lambda$ was varied in the range [0, 1) with a step size of 0.01 ($N = 100$). The hyperparameter $\gamma$ was tuned on the validation set via grid search over $\{0.1, 0.2, 0.4, 0.7, 1\}$. For Pareto front evaluation, we used (1.1, –0.1) as the reference point\footnote{The 2D solution space bounded by $R$ is 1.1 × 1.1 = 1.21. Each HVI (Eq. (\textcolor{red}{12})) is normalized to a percentage: $\text{HVI}(\%) = \frac{\text{HVI}\times10^2}{1.21}$.}, and (0, 1) as the utopia point. We report the mean and variance of all experimental results over three independent runs. All experiments were done using PyTorch and run on two NVIDIA A100 GPUs. Please refer to \textbf{Appendix \textcolor{red}{F}} for more implementation details.


\noindent\textbf{Comparisons with Existing Methods:} We compared our proposed method against several baselines, including four DG methods: DANN \cite{ganin2016domain}, CORAL \cite{sun2016deep}, MMD-AAE \cite{li2018domain}, and DDG \cite{zhang2022towards}. We also evaluated four optimization-based fairness methods capable of producing utility-fairness trade-offs: LNL \cite{kim2019learning}, MaxEnt-ARL \cite{roy2019mitigating}, FairHSIC \cite{quadrianto2019discovering}, and U-FaTE \cite{dehdashtian2024utility}. In addition, we included two recent works targeting the FairDG problem: FEDORA \cite{zhao2024algorithmic} and FATDM-StarGAN \cite{pham2023fairness}. Experiments were conducted on the CelebA, AffectNet, and Jigsaw datasets, evaluating both the Pareto front by HVI (\%) and the selected single solution (Eq. (\textcolor{red}{13})). \textbf{\textit{As DG methods do not consider fairness, they do not yield a Pareto front and are therefore only compared with the single solution.}}

As shown in Table~\textcolor{red}{3}, DG methods achieve high accuracy on the unseen domain but show large fairness violations. Fairness-only methods improve fairness by sacrificing accuracy, but due to poor robustness to domain shifts, their fairness violations remain high compared to FairDG methods. FairDG approaches outperform fairness-only methods by yielding both higher accuracy and lower fairness violations. As shown in Fig.~\textcolor{red}{3}, they also produce higher HVI scores than fairness-only methods. Among FairDG methods, our method consistently achieves the best Pareto front with the highest HVI (\%) across datasets and fairness metrics. It also provides a single solution that dominates existing FairDG baselines, delivering large fairness gains with minimal accuracy loss. To evaluate the benefit of our representation-level encoder design, we implemented a variant Ours-S, which computes conditional dCor using $\mathbf{D}_S$ and $\mathbf{G}$ instead of $\mathbf{Z}_D$ and $\mathbf{Z}_G$. Our method consistently outperforms Ours-S, confirming the advantage of our proposed design.

\begin{figure*}[t]
  \centering
  \includegraphics[height=3.5cm, width=17.7cm]{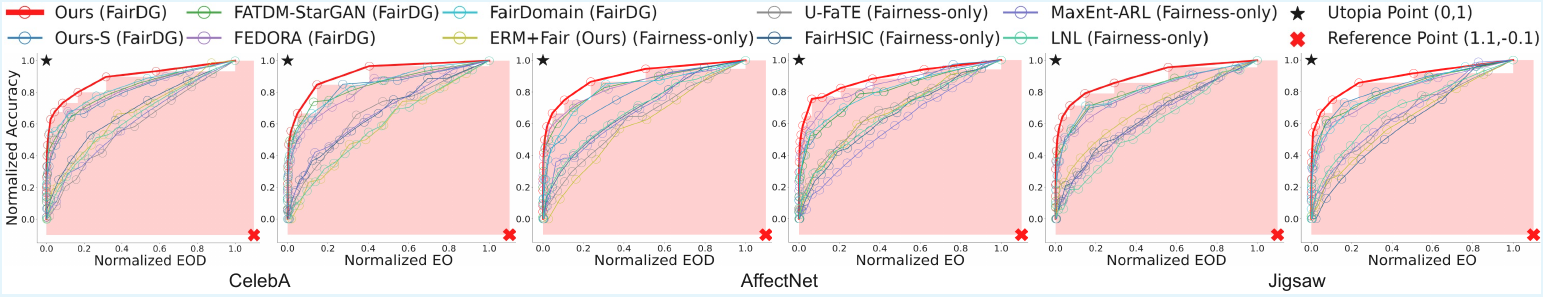}
  \caption{Visualization of Pareto fronts ($\mathcal{P_{\text{norm}}}$) for the fairness-only and FairDG methods. The method with the highest HVI is visualized (the shaded area). Please refer to the \textbf{Appendix \textcolor{red}{G} }for the exact HVI values corresponding to this figure.}
\end{figure*}

\begin{table*}[t]
\centering
\caption{Comparison of our method with different dependence metrics for both $\mathcal{P_{\text{norm}}}$ and $(V_{\text{opt}}, U_{\text{opt}})$ evaluations. }

\setlength{\tabcolsep}{0.8pt} 

\renewcommand{\arraystretch}{1.1} 

{\fontsize{7pt}{9pt}\selectfont
\begin{tabular}{c|cc|cc|cc|ccc|ccc|ccc}
\hline
 Dataset &
  \multicolumn{2}{c|}{CelebA} &
  \multicolumn{2}{c|}{AffectNet} &
  \multicolumn{2}{c|}{Jigsaw} &
  \multicolumn{3}{c|}{CelebA} &
  \multicolumn{3}{c|}{AffectNet} &
  \multicolumn{3}{c}{Jigsaw} \\ \hline
  Methods &
  \multicolumn{1}{c|}{HVI (EOD) \textcolor{red}{$\uparrow$}} & 
  \multicolumn{1}{c|}{HVI (EO) \textcolor{red}{$\uparrow$}} &
  \multicolumn{1}{c|}{HVI (EOD) \textcolor{red}{$\uparrow$}} &
  \multicolumn{1}{c|}{HVI (EO) \textcolor{red}{$\uparrow$}} &
  \multicolumn{1}{c|}{HVI (EOD) \textcolor{red}{$\uparrow$}} &
  \multicolumn{1}{c|}{HVI (EO) \textcolor{red}{$\uparrow$}} &
  \multicolumn{1}{c|}{Acc \textcolor{red}{$\uparrow$}} &
  \multicolumn{1}{c|}{EOD \textcolor{blue}{\textdownarrow}} &
  \multicolumn{1}{c|}{EO \textcolor{blue}{\textdownarrow}} &
  \multicolumn{1}{c|}{Acc \textcolor{red}{$\uparrow$}} &
  \multicolumn{1}{c|}{EOD \textcolor{blue}{\textdownarrow}} &
  \multicolumn{1}{c|}{EO \textcolor{blue}{\textdownarrow}} &
  \multicolumn{1}{c|}{Acc \textcolor{red}{$\uparrow$}} &
  \multicolumn{1}{c|}{EOD \textcolor{blue}{\textdownarrow}} &
  \multicolumn{1}{c}{EO \textcolor{blue}{\textdownarrow}}\\ \hline
\multicolumn{1}{c|}{MINE} & 56.4 ±0.9 & 59.5 ±1.2 & 56.2 ±0.7 & 56.9 ±1.1 & 57.3 ±0.6 & 56.5 ±1.0 & 78.9 ±0.8 & 13.8 ±0.5 & 9.2 ±1.3 & 60.9 ±1.1 & 14.4 ±0.7 & 9.3 ±1.4 & 80.4 ±1.2 & 15.7 ±0.6 & 9.3 ±1.0 \\
\multicolumn{1}{c|}{HSIC} &
  \underline{74.2} ±0.6 & \underline{77.4} ±1.3 & \underline{74.2} ±0.5 & \underline{71.4} ±1.2 & \underline{72.3} ±0.7 & \underline{73.5} ±0.9 & \underline{86.8} ±1.0 & \underline{8.4} ±0.6 &
  \underline{3.9} ±1.1 & \underline{66.8} ±0.8 & \underline{8.6} ±1.4 &
  \underline{6.3} ±0.9 & \underline{85.3} ±1.2 & \underline{10.2} ±0.5 & \underline{4.7} ±1.3 \\
\multicolumn{1}{c|}{dCor} &
  \textbf{\underline{75.4}} ±1.1 & \textbf{\underline{78.3}} ±0.7 & \textbf{\underline{76.4}} ±0.8 & \textbf{\underline{74.9}} ±1.4 & \textbf{\underline{75.8}} ±0.9 & \textbf{\underline{75.7}} ±1.2 & \textbf{\underline{88.7}} ±1.3 & \textbf{\underline{8.2}} ±0.6 &
  \textbf{\underline{3.1}} ±0.7 & \textbf{\underline{68.9}} ±0.5 & \textbf{\underline{8.4}} ±0.8 &
  \textbf{\underline{4.3}} ±1.1 & \underline{86.3} ±0.9 & \textbf{\underline{9.7}} ±1.4 & \textbf{\underline{3.7}} ±0.6 \\ \hline
\end{tabular}
}
\end{table*}

\begin{table}[t]
\centering
\caption{Comparisons of our proposed method with different numbers of source domains using the AffectNet dataset.}

\setlength{\tabcolsep}{2.5pt} 

\renewcommand{\arraystretch}{1.1} 

{\fontsize{7pt}{9pt}\selectfont
\begin{tabular}{c|cc|ccc}
\hline
  Methods &
  \multicolumn{1}{c|}{HVI (EOD) \textcolor{red}{$\uparrow$}} &
  \multicolumn{1}{c|}{HVI (EO) \textcolor{red}{$\uparrow$}} &
  \multicolumn{1}{c|}{Acc \textcolor{red}{$\uparrow$}} &
  \multicolumn{1}{c|}{EOD \textcolor{blue}{\textdownarrow}} &
  \multicolumn{1}{c}{EO \textcolor{blue}{\textdownarrow}}\\ \hline
\multicolumn{1}{c|}{\textbf{Ours} (2 domains)} & 75.3 ±0.6 & 73.5 ±1.0 & 67.8 ±1.1 & 8.8 ±0.5 & 5.2 ±1.3 \\
\multicolumn{1}{c|}{\textbf{Ours} (3 domains)} &
 \underline{76.4} ±0.7 & \underline{74.9} ±0.9 & \underline{68.9} ±0.8 & \underline{8.4} ±1.1 &
  \underline{4.3} ±0.6 \\
\multicolumn{1}{c|}{\textbf{Ours} (5 domains)} &
   \textbf{\underline{77.3}} ±0.8 & \textbf{\underline{75.5}} ±1.2 & \textbf{\underline{69.7}} ±0.9 & \textbf{\underline{8.1}} ±0.7 & \textbf{\underline{3.9}} ±1.0 \\ \hline
\end{tabular}
}
\end{table}

\begin{table}[t]
\centering
\caption{Comparisons of the existing FairDG methods with a different domain split using the AffectNet dataset.}

\setlength{\tabcolsep}{2.4pt} 

\renewcommand{\arraystretch}{1.1} 

{\fontsize{7pt}{9pt}\selectfont
\begin{tabular}{c|cc|ccc}
\hline
  Methods &
  \multicolumn{1}{c|}{HVI (EOD) \textcolor{red}{$\uparrow$}} &
  \multicolumn{1}{c|}{HVI (EO) \textcolor{red}{$\uparrow$}} &
  \multicolumn{1}{c|}{Acc \textcolor{red}{$\uparrow$}} &
  \multicolumn{1}{c|}{EOD \textcolor{blue}{\textdownarrow}} &
  \multicolumn{1}{c}{EO \textcolor{blue}{\textdownarrow}}\\ \hline
\multicolumn{1}{c|}{\textcolor{red}{$\dagger*$} FEDORA} & 76.6 ±0.2 & 73.1 ±0.4 & 66.7 ±0.6 & 8.5 ±1.0 & \underline{4.5} ±0.6 \\
\multicolumn{1}{c|}{\textcolor{red}{$\dagger*$} FATDM-StarGAN} & \underline{77.9} ±0.8 & \underline{76.7} ±0.8 & \underline{67.9} ±0.6 & \underline{7.5} ±1.1 & 4.9 ±0.4 \\
\multicolumn{1}{c|}{\textcolor{red}{$\dagger*$} \textbf{Ours}} & \textbf{\underline{79.3}} ±0.4 & \textbf{\underline{77.1}} ±0.1 & \textbf{\underline{68.1}} ±0.3 & \textbf{\underline{6.0}} ±0.6 & \textbf{\underline{4.1}} ±0.5 \\ \hline
\end{tabular}
}
\end{table}

\noindent\textbf{Ablation Studies:} As shown in Table \textcolor{red}{3}, we ablate different components of Eq. (\textcolor{red}{11}): ERM alone, ERM with the fairness constraint (ERM+Fair), and ERM with the source domain invariance constraint (ERM+SDI). Compared to ERM, ERM+SDI significantly improves target domain accuracy on both datasets, confirming the benefit of enforcing domain invariance. ERM+Fair lowers EO and EOD violations compared to ERM, demonstrating the effectiveness of the fairness constraint, though at the cost of accuracy. The full optimization of Eq. (\textcolor{red}{11}) achieves the best trade-off between utility and fairness in the target domain.

\noindent\textbf{Different Dependence Metrics:} As shown in Table \textcolor{red}{4}, dCor consistently outperforms HSIC and MINE across all settings. This is because MINE \cite{pmlr-v80-belghazi18a} only approximates the lower bound of MI through the Donsker-Varadhan (DV) representation. Minimizing the lower bound, however, does not guarantee that MI is minimized. Moreover, MINE introduces an inherent approximation error in addition to stochastic error that could be reduced as the sample size increases, whereas dCor and HSIC are only subject to stochastic error. However, HSIC is sensitive to kernel selection and parameter tuning. In contrast, dCor is parameter-free and operates directly on pairwise distances in the original data space, making it robust against kernel distortions.

\noindent\textbf{Impact of the number of source domains:} Like other DG methods, our method relies on a set of source domains to generalize to an unseen target domain. To discuss the effect of the number of source domains during training, we split the source domains (perceived age groups) in the AffectNet dataset into two and five groups: {30--59, 60--70+} and {30--39, 40--49, 50--59, 60--69, 70+}, respectively, in addition to the original setup with three source domains {30--49, 50--69, 70+}. The validation and test domains remain unchanged. As shown in Table~\textcolor{red}{5}, we observe that increasing the number of source domain splits improves the trade-off achieved by our method. However, splitting domains at a finer granularity may be difficult and require extra human effort.

\noindent\textbf{Robustness of domains splits:} In addition to evaluating our method across datasets of different modalities, we further assess its robustness to varying domain splits within the same dataset. We run an experiment on the AffectNet dataset, where the unseen test domain is instead set to the 70+ age group, and the remaining age groups are used for training and validation. As shown in Table~\textcolor{red}{6}, our method still consistently outperforms existing FairDG approaches in both the Pareto front and single-solution under this setting.

\section{Conclusion}

In this paper, we study the FairDG problem, which aims to minimize both expected risk and fairness violations in unseen target domains. We derive novel upper bounds based on MIs for both the expected risk and fairness violations in multi-class classification tasks with multi-group sensitive attributes, offering key insights from an information-theoretic perspective that inform algorithm design. Guided by these insights, we introduce a practical framework that models the utility-fairness trade-off through Pareto optimization. Experimental results on real-world natural language and vision datasets show that our method outperforms existing approaches, achieving better utility-fairness trade-offs.

\noindent\textbf{Ethical Statement.} The sensitive attributes used in this work are either perceived or purely linguistic in meaning. They do not represent any individual’s self-identification.

\section{Acknowledgments}

Tangzheng Lian is fully funded by the faculty of Natural, Mathematics \& Engineering Sciences (NMES) PhD studentship at King's College London.

\bibliography{aaai2026}

\newpage
\onecolumn 
\begin{center}
    {\LARGE \textbf{Fair Domain Generalization: An Information-Theoretic View}} \\[0.5cm] 
    {\LARGE Supplementary Material}
\end{center}
\vspace{1cm}

\noindent\textbf{\Large  Overview}\\

\begin{enumerate}
    \item \textbf{Appendix \textcolor{red}{A}}: Supporting lemmas for the proofs of the theorems.
    \item \textbf{Appendix \textcolor{red}{B}}: Proofs of the theorems in the manuscript.
    \item \textbf{Appendix \textcolor{red}{C}}:  Proofs of the supporting lemmas.
    \item \textbf{Appendix \textcolor{red}{D}}: Comparisons with the previous bounds in the FairDG problem.
    \item \textbf{Appendix \textcolor{red}{E}}: Calculations of empirical distance correlations (Eq. (\textcolor{red}{8}) and Eq. (\textcolor{red}{9}) in the manuscript).
    \item \textbf{Appendix \textcolor{red}{F}}: Datasets and the implementation details.
    \item \textbf{Appendix \textcolor{red}{G}}: Table \textcolor{red}{4} with variances and the corresponding HVI values of the Pareto fronts in Figure \textcolor{red}{3}.\\
\end{enumerate} 

\noindent\textbf{\Large  \textcolor{red}{A}. Lemmas}\\

\noindent\textbf{Lemma 1.}  
Let \( \mathbf{X} \) be a random variable (continuous or discrete) supported on \( \mathcal{X} \subseteq \mathbb{R}^n \), and let \( f: \mathcal{X} \rightarrow [0, C] \) be a measurable function bounded above by a constant \( C > 0 \). Consider two domains \( d_j \) and \( d_i \) with probability distributions $P_{d_j}^{\mathbf{X}}$ and $P_{d_i}^{\mathbf{X}}$ over \( \mathbf{X} \). Then, the following inequality holds:
\[
\mathbb{E}_{d_j}[f(\mathbf{X})] - \mathbb{E}_{d_i}[f(\mathbf{X})] \leq C \cdot \delta_{\mathrm{TV}}(P_{d_j}^{\mathbf{X}}, P_{d_i}^{\mathbf{X}}).
\]

\noindent\textbf{Lemma 2.}  
Let \( \mathbf{X} \) be a discrete random variable taking values in a finite set \( \mathcal{X} \), and let \( \mathbf{Y} \) be a random variable (discrete or continuous) supported on \( \mathcal{Y} \subseteq \mathbb{R}^n \). Let $p(x)$ denote the probability mass function of $\mathbf{X}$ and let $P_{\mathbf{Y}}$ denote the marginal distribution of $\mathbf{Y}$. Denote by $P_{\mathbf{Y}|x}$ the conditional distribution of $\mathbf{Y}$ given $\mathbf{X} = x$. Then, the following inequality holds:
\[
\sum_{x \in \mathcal{X}} p(x)\, \delta_{\mathrm{TV}}\!\Bigl(P_{\mathbf{Y}|x},\, P_\mathbf{Y}\Bigr) 
\;\leq\; 
\sqrt{\frac{I(\mathbf{X}; \mathbf{Y})}{2}}.
\]

\noindent\textbf{Lemma 3.}  
Consider two domains \( d_i \) and \( d_j \). Let \( P_{d_i} \) and \( P_{d_i}' \) be two probability distributions defined under domain \( d_i \), and let \( P_{d_j} \) and \( P_{d_j}' \) be two distributions under domain \( d_j \). Then, the following inequality holds:
\[
\delta_{\mathrm{TV}}\bigl(P_{d_j},\, P_{d_j}'\bigr) - \delta_{\mathrm{TV}}\bigl(P_{d_i},\, P_{d_i}'\bigr) 
\leq 
\delta_{\mathrm{TV}}\bigl(P_{d_j},\, P_{d_i}\bigr) + \delta_{\mathrm{TV}}\bigl(P_{d_j}',\, P_{d_i}'\bigr).
\]

\noindent\textbf{Lemma 4.}  
Let \( \mathbf{X}, \mathbf{D}, \mathbf{G} \) be discrete random variables with finite ranges \( \mathcal{X}, \mathcal{D}, \mathcal{G} \), respectively, and let \( \mathbf{Y} \) be a random variable supported on \( \mathcal{Y} \subseteq \mathbb{R}^n \), which may be either discrete or continuous. Let
\(
p(x, d, g)
\) denote the joint probability mass function over $\mathcal{X} \times \mathcal{D} \times \mathcal{G}$,
and let \( P_{\mathbf{Y} \mid x, d, g} \) and \( P_{\mathbf{Y} \mid d, g} \) denote the conditional distributions of \( \mathbf{Y} \) given \((x,\,d,\, g) \) and \( (d,\,g) \), respectively. Then, the following inequality holds:
\[
\sum_{x \in \mathcal{X}}\sum_{d \in \mathcal{D}} \sum_{g \in \mathcal{G}}
p(x, d, g)\,
\delta_{\mathrm{TV}}\bigl(P_{\mathbf{Y} \mid x, d, g},\, P_{\mathbf{Y} \mid d, g}\bigr)
\;\le\;
\sqrt{\frac{I(\mathbf{X}; \mathbf{Y} \mid \mathbf{D}, \mathbf{G})}{2}\, }.
\] \\

\noindent\textbf{\Large  \textcolor{red}{B}. Proofs of Theorems}\\

\noindent\textbf{Proof of Theorem 1.}  
Applying \textbf{Lemma 1} with the substitutions \( f \to \mathcal{L}(\cdot) \), \( \mathbf{X} \to (\hat{f}_\theta(\mathbf{X}), \mathbf{Y}) \), \( d_j \to d_T \), and \( d_i \to d_S \), we obtain:

\[
\mathcal{R}_{d_T}(\hat{f}_\theta) \leq \mathcal{R}_{d_S}(\hat{f}_\theta) + C \cdot \delta_{\mathrm{TV}}\Bigl(P^{\hat{f}_\theta(\mathbf{X}), \mathbf{Y}}_{d_T},\, P^{\hat{f}_\theta(\mathbf{X}), \mathbf{Y}}_{d_S}\Bigr).
\]

By the triangle inequality \( \delta_{\mathrm{TV}}(P, Q) \leq \delta_{\mathrm{TV}}(P, M) + \delta_{\mathrm{TV}}(M, Q) \) and the symmetry of TV distance \( \delta_{\mathrm{TV}}(M, Q) = \delta_{\mathrm{TV}}(Q, M)\), we have:

\[
\delta_{\mathrm{TV}}\Bigl(P^{\hat{f}_\theta(\mathbf{X}),\, \mathbf{Y}}_{d_T},\, P^{\hat{f}_\theta(\mathbf{X}),\, \mathbf{Y}}_{d_S}\Bigr) 
\leq 
\delta_{\mathrm{TV}}\Bigl(P^{\hat{f}_\theta(\mathbf{X}),\, \mathbf{Y}}_{d_T},\, P^{\hat{f}_\theta(\mathbf{X}),\, \mathbf{Y}}\Bigr) + \delta_{\mathrm{TV}}\Bigl(P^{\hat{f}_\theta(\mathbf{X}),\, \mathbf{Y}}_{d_S},\, P^{\hat{f}_\theta(\mathbf{X}),\, \mathbf{Y}}\Bigr),
\]

\noindent where we set \( P = P^{\hat{f}_\theta(\mathbf{X}),\, \mathbf{Y}}_{d_T} \), \( Q = P^{\hat{f}_\theta(\mathbf{X}),\, \mathbf{Y}}_{d_S} \), and \( M = P^{\hat{f}_\theta(\mathbf{X}),\, \mathbf{Y}} \). Here, 
\(P^{\hat f_\theta(\mathbf X),\,\mathbf Y}
\;=\;
\sum_{d_S \in \mathcal{D}_S} p(d_S)\;P^{\hat f_\theta(\mathbf X),\,\mathbf Y}_{d_S}\) is a mixture distribution over the source domains. Substituting this into the previous inequality yields:
\[
\mathcal{R}_{d_T}(\hat{f}_\theta) 
\leq 
\mathcal{R}_{d_S}(\hat{f}_\theta) + C \cdot \delta_{\mathrm{TV}}\Bigl(P^{\hat{f}_\theta(\mathbf{X}),\, \mathbf{Y}}_{d_T},\, P^{\hat{f}_\theta(\mathbf{X}),\, \mathbf{Y}}\Bigr) + C \cdot \delta_{\mathrm{TV}}\Bigl(P^{\hat{f}_\theta(\mathbf{X}),\, \mathbf{Y}}_{d_S},\, P^{\hat{f}_\theta(\mathbf{X}),\, \mathbf{Y}}\Bigr).
\]

Multiplying both sides by \( p(d_S) \) and summing over \( d_S \in \mathcal{D}_S \), we then obtain:
\[
\begin{aligned}
\sum_{d_S \in \mathcal{D}_S} p(d_S)\, \mathcal{R}_{d_T}(\hat{f}_\theta) \leq \sum_{d_S \in \mathcal{D}_S} p(d_S)\, \mathcal{R}_{d_S}(\hat{f}_\theta) &+ C \cdot \sum_{d_S \in \mathcal{D}_S} p(d_S)\delta_{\mathrm{TV}}\Bigl(P^{\hat{f}_\theta(\mathbf{X}), \mathbf{Y}}_{d_T},\, P^{\hat{f}_\theta(\mathbf{X}), \mathbf{Y}}\Bigr) \\ &\hspace{4cm}+ C \cdot \sum_{d_S \in \mathcal{D}_S} p(d_S)\delta_{\mathrm{TV}}\Bigl(P^{\hat{f}_\theta(\mathbf{X}), \mathbf{Y}}_{d_S},\, P^{\hat{f}_\theta(\mathbf{X}), \mathbf{Y}}\Bigr).
\end{aligned}
\]

Since \( \sum_{d_S \in \mathcal{D}_S} p(d_S) = 1 \), this simplifies to:
\[
\mathcal{R}_{d_T}(\hat{f}_\theta) \leq \underbrace{\sum_{d_S \in \mathcal{D}_S} p(d_S)\, \mathcal{R}_{d_S}(\hat{f}_\theta)}_{\textbf{Term (1)}} + C \cdot \underbrace{\delta_{\mathrm{TV}}\Bigl(P^{\hat{f}_\theta(\mathbf{X}), \mathbf{Y}}_{d_T},\, P^{\hat{f}_\theta(\mathbf{X}), \mathbf{Y}}\Bigr)}_{\textbf{Term (2)}} + C \underbrace{\cdot\sum_{d_S \in \mathcal{D}_S} p(d_S)\,  \delta_{\mathrm{TV}}\Bigl(P^{\hat{f}_\theta(\mathbf{X}), \mathbf{Y}}_{d_S},\, P^{\hat{f}_\theta(\mathbf{X}), \mathbf{Y}}\Bigr)}_{\textbf{Term (3)}}.
\]

For \textbf{Term (1)}, using the tower rule of expectation, we have:
\[
\begin{aligned}
\sum_{d_S \in \mathcal{D}_S} p(d_S)\, \mathcal{R}_{d_S}(\hat{f}_\theta) &= \sum_{d_S\in\mathcal{D}_S}
p(d_S)\,
\mathbb{E}_{(\mathbf{X},\mathbf{Y})\sim d_S}\bigl[\,\mathcal L(\hat f_\theta(\mathbf{X}),\mathbf{Y})\bigr] \\&= \mathbb{E}_{d_S\sim \mathcal{D}_S}\bigl[\mathbb{E}_{(\mathbf{X},\mathbf{Y})\sim d_S}\bigl[\,\mathcal L(\hat f_\theta(\mathbf{X}),\mathbf{Y})\bigr]\bigr] \\&= \mathbb{E}_{(\mathbf{X},\mathbf{Y})\sim \mathcal{D}_S}\bigl[\,\mathcal L(\hat f_\theta(\mathbf{X}),\mathbf{Y})\\&= \mathcal{R}_{\mathcal{D}_S}(\hat{f}_\theta)
\end{aligned}
\]

For \textbf{Term (2)}, using the data–processing inequality (DPI) for $f$–divergences (i.e., total variance distance), we have:

\[\;
\delta_{\mathrm{TV}}\!\Bigl(P^{\hat{f}_\theta(\mathbf{X}),\mathbf{Y}}_{d_T},\;P^{\hat{f}_\theta(\mathbf{X}),\mathbf{Y}}\Bigr)
\;\le\;
\delta_{\mathrm{TV}}\!\Bigl(P^{\mathbf{X},\mathbf{Y}}_{d_T},\;P^{\mathbf{X},\mathbf{Y}}\Bigr).
\;
\]

For \textbf{Term (3)}, applying \textbf{Lemma 2} (with the substitution \( \mathbf{X} \to \mathbf{D}_S \) and \( \mathbf{Y} \to (\hat{f}_\theta(\mathbf{X}), \mathbf{Y}) \), we have:
\[
\sum_{d_S \in \mathcal{D}_S} p(d_S)\, \delta_{\mathrm{TV}}\Bigl(P^{\hat{f}_\theta(\mathbf{X}), \mathbf{Y}}_{d_S} , P^{\hat{f}_\theta(\mathbf{X}), \mathbf{Y}}\Bigr) \leq \sqrt{\frac{I\Bigl(\mathbf{D}_S;\, (\hat{f}_\theta(\mathbf{X}), \mathbf{Y})\Bigr)}{2}}.
\]

Thus, we obtain the desired bound:
\[
\mathcal{R}_{d_T}(\hat{f}_\theta) \leq \mathcal{R}_{\mathcal{D}_S}(\hat{f}_\theta) + C \cdot \delta_{\mathrm{TV}}\Bigl(P^{\mathbf{X}, \mathbf{Y}}_{d_T},\, P^{\mathbf{X}, \mathbf{Y}}\Bigr) + \frac{\sqrt{2}C}{2}\sqrt{I\Bigl(\mathbf{D}_S;\, (\hat{f}_\theta(\mathbf{X}), \mathbf{Y})\Bigr)}.
\]

\qed \\

\noindent\textbf{Proof of Theorem 2.} The difference in EOD violations between the target domain \( d_T \) and a source domain \( d_S \) is given by:
\[
\Delta_{d_T}^{\mathrm{EOD}}(\hat{f}_\theta) - \Delta_{d_S}^{\mathrm{EOD}}(\hat{f}_\theta) = \frac{2}{|\mathcal{Y}||\mathcal{G}|(|\mathcal{G}| - 1)} \sum_{y \in \mathcal{Y}} \sum_{\{g, g'\} \subset \mathcal{G}} \delta_{\mathrm{TV}}\bigl(P^{\hat f_\theta(\mathbf{X}) \mid y,g}_{d_T},\, P^{\hat f_\theta(\mathbf{X}) \mid y,g'}_{d_T}\bigr) - \delta_{\mathrm{TV}}\bigl(P^{\hat f_\theta(\mathbf{X}) \mid y,g}_{d_S},\, P^{\hat f_\theta(\mathbf{X}) \mid y,g'}_{d_S}\bigr).
\]

Applying \textbf{Lemma 3} with the identifications \( P_{d_j} \mapsto P^{\hat f_\theta(\mathbf{X}) \mid y,g}_{d_T}\), \( P_{d_j}' \mapsto P^{\hat f_\theta(\mathbf{X}) \mid y,g'}_{d_T} \), \( P_{d_i} \mapsto P^{\hat f_\theta(\mathbf{X}) \mid y,g}_{d_S} \), and \( P_{d_i}' \mapsto P^{\hat f_\theta(\mathbf{X}) \mid y,g'}_{d_S} \), we obtain:
\[
\delta_{\mathrm{TV}}\bigl(P^{\hat f_\theta(\mathbf{X}) \mid y,g}_{d_T}, P^{\hat f_\theta(\mathbf{X}) \mid y,g'}_{d_T}\bigr) - \delta_{\mathrm{TV}}\bigl(P^{\hat f_\theta(\mathbf{X}) \mid y,g}_{d_S}, P^{\hat f_\theta(\mathbf{X}) \mid y,g'}_{d_S}\bigr) \leq \delta_{\mathrm{TV}}\bigl(P^{\hat f_\theta(\mathbf{X}) \mid y,g}_{d_T}, P^{\hat f_\theta(\mathbf{X}) \mid y,g}_{d_S}\bigr) + \delta_{\mathrm{TV}}\bigl(P^{\hat f_\theta(\mathbf{X}) \mid y,g'}_{d_T}, P^{\hat f_\theta(\mathbf{X}) \mid y,g'}_{d_S}\bigr).
\]

Substituting back and simplifying the sum over pairs \( \{g, g'\} \), we get:
\[
\begin{aligned}
\Delta_{d_T}^{\mathrm{EOD}} - \Delta_{d_S}^{\mathrm{EOD}} &\leq  \frac{2}{|\mathcal{Y}||\mathcal{G}|(|\mathcal{G}| - 1)} \sum_{y \in \mathcal{Y}} \sum_{\{g, g'\} \subset \mathcal{G}} 
\left[\delta_{\mathrm{TV}}\bigl(P^{\hat f_\theta(\mathbf{X}) \mid y,g}_{d_T},\, P^{\hat f_\theta(\mathbf{X}) \mid y,g}_{d_S}\bigr) + 
\delta_{\mathrm{TV}}\bigl(P^{\hat f_\theta(\mathbf{X}) \mid y,g'}_{d_T},\, P^{\hat f_\theta(\mathbf{X}) \mid y,g'}_{d_S}\bigr)\right] \\&= \frac{2}{|\mathcal{Y}||\mathcal{G}|(|\mathcal{G}| - 1)} \cdot (|\mathcal{G}|-1) \sum_{y \in \mathcal{Y}} \sum_{g \in \mathcal{G}} 
\delta_{\mathrm{TV}}\bigl(P^{\hat f_\theta(\mathbf{X}) \mid y,g}_{d_T},\, P^{\hat f_\theta(\mathbf{X}) \mid y,g}_{d_S}\bigr) \\&= \frac{2}{|\mathcal{Y}||\mathcal{G}|} \sum_{y \in \mathcal{Y}} \sum_{g \in \mathcal{G}} 
\delta_{\mathrm{TV}}\bigl(P^{\hat f_\theta(\mathbf{X}) \mid y,g}_{d_T},\, P^{\hat f_\theta(\mathbf{X}) \mid y,g}_{d_S}\bigr).
\end{aligned}
\]

By the triangle inequality \( \delta_{\mathrm{TV}}(P, Q) \leq \delta_{\mathrm{TV}}(P, M) + \delta_{\mathrm{TV}}(M, Q) \) and the symmetry of TV distance \( \delta_{\mathrm{TV}}(M, Q) = \delta_{\mathrm{TV}}(Q, M)\), we have:

\[
\delta_{\mathrm{TV}}\bigl(P^{\hat f_\theta(\mathbf{X}) \mid y,g}_{d_T},\, P^{\hat f_\theta(\mathbf{X}) \mid y,g}_{d_S}\bigr) \leq 
\delta_{\mathrm{TV}}\bigl(P^{\hat f_\theta(\mathbf{X}) \mid y,g}_{d_T},\, P^{\hat f_\theta(\mathbf{X}) \mid y,g}\bigr) + \delta_{\mathrm{TV}}\bigl(P^{\hat f_\theta(\mathbf{X}) \mid y,g}_{d_S},\, P^{\hat f_\theta(\mathbf{X}) \mid y,g}\bigr)
\]

\noindent where we set \( P = P^{\hat f_\theta(\mathbf{X}) \mid y,g}_{d_T} \), \( Q = P^{\hat f_\theta(\mathbf{X}) \mid y,g}_{d_S} \), and \( M = P^{\hat f_\theta(\mathbf{X}) \mid y,g} \). Here, 
\(P^{\hat f_\theta(\mathbf{X}) \mid y,g}
\;=\;
\sum_{d_S \in \mathcal{D}_S} p(d_S)\;P^{\hat f_\theta(\mathbf{X}) \mid y,g}_{d_S}\) is a mixture distribution over the source domains. Substituting this into the previous inequality yields:
\[
\Delta_{d_T}^{\mathrm{EOD}} (\hat{f}_\theta) 
\leq \Delta_{d_S}^{\mathrm{EOD}}(\hat{f}_\theta)+\frac{2}{|\mathcal{Y}||\mathcal{G}|} \sum_{y \in \mathcal{Y}} \sum_{g \in \mathcal{G}} 
\left[\delta_{\mathrm{TV}}\bigl(P^{\hat f_\theta(\mathbf{X}) \mid y,g}_{d_T},\, P^{\hat f_\theta(\mathbf{X}) \mid y,g}\bigr) + \delta_{\mathrm{TV}}\bigl(P^{\hat f_\theta(\mathbf{X}) \mid y,g}_{d_S},\, P^{\hat f_\theta(\mathbf{X}) \mid y,g}\bigr)\right]
\]

Multiplying both sides by by \( p(d_S|y,g) \), and sum over all \( d_S \in \mathcal{D}_S \):
\[
\begin{aligned}
\sum_{d_S \in \mathcal{D}_S}p(d_S|y,g) \Delta_{d_T}^{\mathrm{EOD}}(\hat{f}_\theta) &\leq \sum_{d_S \in \mathcal{D}_S}p(d_S|y,g) \Delta_{d_S}^{\mathrm{EOD}}(\hat{f}_\theta) \\ 
&+ \frac{2}{|\mathcal{Y}||\mathcal{G}|} \sum_{d_S \in \mathcal{D}_S}p(d_S|y,g) \sum_{y \in \mathcal{Y}} \sum_{g \in \mathcal{G}} 
\left[\delta_{\mathrm{TV}}\bigl(P^{\hat f_\theta(\mathbf{X}) \mid y,g}_{d_T}, P^{\hat f_\theta(\mathbf{X}) \mid y,g}\bigr) + \delta_{\mathrm{TV}}\bigl(P^{\hat f_\theta(\mathbf{X}) \mid y,g}_{d_S},\, P^{\hat f_\theta(\mathbf{X}) \mid y,g}\bigr)\right]
\end{aligned}
\]

Since $\sum_{d_S \in \mathcal{D}_S}p(d_S|y,g) = 1$, we have:
\begin{equation}\label{eqn:einstein}
\begin{aligned}
 \Delta_{d_T}^{\mathrm{EOD}}(\hat{f}_\theta) \leq \underbrace{\sum_{d_S \in \mathcal{D}_S}p(d_S|y,g) \Delta_{d_S}^{\mathrm{EOD}}(\hat{f}_\theta)}_{\textbf{Term (1)}} &+ \underbrace{\frac{2}{|\mathcal{Y}||\mathcal{G}|} \sum_{y \in \mathcal{Y}} \sum_{g \in \mathcal{G}} 
\delta_{\mathrm{TV}}\bigl(P^{\hat f_\theta(\mathbf{X}) \mid y,g}_{d_T},\, P^{\hat f_\theta(\mathbf{X}) \mid y,g}\bigr)}_{\textbf{Term (2)}} \\&+ \underbrace{\frac{2}{|\mathcal{Y}||\mathcal{G}|} \sum_{d_S \in \mathcal{D}_S}\sum_{y \in \mathcal{Y}} \sum_{g \in \mathcal{G}}p(d_S|y,g)\delta_{\mathrm{TV}}\bigl(P^{\hat f_\theta(\mathbf{X}) \mid y,g}_{d_S},\, P^{\hat f_\theta(\mathbf{X}) \mid y,g}\bigr)}_{\textbf{Term (3)}}.
\end{aligned}
\tag{\textcolor{red}{$\dagger$}}
\end{equation}

For the \textbf{Term (1)}, we have:

\[
\textbf{Term (1)} = \frac{2}{|\mathcal{Y}||\mathcal{G}|(|\mathcal{G}| - 1)} \sum_{d_S \in \mathcal{D}_S}p(d_S|y,g)\sum_{y \in \mathcal{Y}} \sum_{\{g, g'\} \subset \mathcal{G}}  \delta_{\mathrm{TV}}\bigl(P^{\hat f_\theta(\mathbf{X}) \mid y,g}_{d_S},\, P^{\hat f_\theta(\mathbf{X}) \mid y,g'}_{d_S}\bigr) 
\]

By the triangle inequality \( \delta_{\mathrm{TV}}(P, Q) \leq \delta_{\mathrm{TV}}(P, M) + \delta_{\mathrm{TV}}(M, Q) \) and the symmetry of TV distance \( \delta_{\mathrm{TV}}(M, Q) = \delta_{\mathrm{TV}}(Q, M)\), we have:

\[
\delta_{\mathrm{TV}}\bigl(P^{\hat f_\theta(\mathbf{X}) \mid y,g}_{d_S},\, P^{\hat f_\theta(\mathbf{X}) \mid y,g'}_{d_S}\bigr) \leq \delta_{\mathrm{TV}}\bigl(P^{\hat f_\theta(\mathbf{X}) \mid y,g}_{d_S},\, P^{\hat f_\theta(\mathbf{X}) \mid y}_{d_S}\bigr)  + \delta_{\mathrm{TV}}\bigl(P^{\hat f_\theta(\mathbf{X}) \mid y,g'}_{d_S},\, P^{\hat f_\theta(\mathbf{X}) \mid y}_{d_S}\bigr)
\]

\noindent where we set \( P = P^{\hat f_\theta(\mathbf{X}) \mid y,g}_{d_S}\), \( Q = P^{\hat f_\theta(\mathbf{X}) \mid y,g'}_{d_S}\), and \( M = P^{\hat f_\theta(\mathbf{X}) \mid y}_{d_S}\). Substituting this into the previous equation yields:

\[
\begin{aligned}
\textbf{Term (1)} &\leq \frac{2}{|\mathcal{Y}||\mathcal{G}|(|\mathcal{G}| - 1)}\sum_{d_S \in \mathcal{D}_S}p(d_S|y,g)\sum_{y \in \mathcal{Y}} \sum_{\{g, g'\} \subset \mathcal{G}}\left[  \delta_{\mathrm{TV}}\bigl(P^{\hat f_\theta(\mathbf{X}) \mid y,g}_{d_S}, P^{\hat f_\theta(\mathbf{X}) \mid y}_{d_S}\bigr)  + \delta_{\mathrm{TV}}\bigl(P^{\hat f_\theta(\mathbf{X}) \mid y,g'}_{d_S}, P^{\hat f_\theta(\mathbf{X}) \mid y}_{d_S}\bigr)\right] \\&=  \frac{2}{|\mathcal{Y}||\mathcal{G}|(|\mathcal{G}| - 1)} \cdot (|\mathcal{G}|-1)\sum_{d_S \in \mathcal{D}_S}p(d_S|y,g)\sum_{y \in \mathcal{Y}} \sum_{g \in \mathcal{G}}  \delta_{\mathrm{TV}}\bigl(P^{\hat f_\theta(\mathbf{X}) \mid y,g}_{d_S},\, P^{\hat f_\theta(\mathbf{X}) \mid y}_{d_S}\bigr) \\&= \frac{2}{|\mathcal{Y}||\mathcal{G}|} \sum_{d_S \in \mathcal{D}_S}p(d_S|y,g)\sum_{y \in \mathcal{Y}} \sum_{g \in \mathcal{G}}  \delta_{\mathrm{TV}}\bigl(P^{\hat f_\theta(\mathbf{X}) \mid y,g}_{d_S},\, P^{\hat f_\theta(\mathbf{X}) \mid y}_{d_S}\bigr) \\&= \frac{2}{|\mathcal{Y}||\mathcal{G}|}\sum_{d_S \in \mathcal{D}_S}\sum_{y \in \mathcal{Y}} \sum_{g \in \mathcal{G}}
\frac{p(d_S,y,g)}{p(y, g)}
\;\delta_{\mathrm{TV}}\bigl(P^{\hat f_\theta(\mathbf{X}) \mid y,g}_{d_S},\,P^{\hat f_\theta(\mathbf{X}) \mid y}_{d_S}\bigr) \\&\leq \frac{2}{|\mathcal{Y}||\mathcal{G}|\displaystyle \min_{\substack{y, g}}p(y,g)}
\sum_{d_S \in \mathcal{D}_S}\sum_{y \in \mathcal{Y}} \sum_{g \in \mathcal{G}}p(d_S,y,g)\,
\delta_{\rm TV}\bigl(P^{\hat f_\theta(\mathbf{X}) \mid y,g}_{d_S},\,P^{\hat f_\theta(\mathbf{X}) \mid y}_{d_S}\bigr).
\end{aligned}
\]

By Lemma 4, with the identification 
\(\mathbf X\!\mapsto\!\mathbf G\), 
\(\mathbf Y\!\mapsto\!\hat f_\theta(\mathbf{X})\), 
\(\mathbf D\!\mapsto\!\mathbf Y\), 
and \(\mathbf G\!\mapsto\!\mathbf D_S\),  
one has
\[
\sum_{d_S \in \mathcal{D}_S}\sum_{y \in \mathcal{Y}} \sum_{g \in \mathcal{G}}p(d_S,y,g)\,
\delta_{\rm TV}\bigl(P^{\hat f_\theta(\mathbf{X}) \mid y,g}_{d_S},\,P^{\hat f_\theta(\mathbf{X}) \mid y}_{d_S}\bigr)
\;\le\;
\sqrt{\frac{I\bigl(\mathbf G;\,\hat f_\theta(\mathbf{X})\mid\mathbf Y,\mathbf D_S\bigr)}{2}}.
\]

Thus, we have the upper bound for \textbf{Term (1)}:

\[
\textbf{Term (1)} \leq \frac{\sqrt{2I\bigl(\mathbf G;\,\hat f_\theta(\mathbf{X})\mid\mathbf Y,\mathbf D_S\bigr)}}{|\mathcal{Y}||\mathcal{G}|\displaystyle \min_{\substack{y, g}}p(y,g)}.
\]

For \textbf{Term (2)}, using the data–processing inequality (DPI) for $f$–divergences (i.e., total variance distance), we have:

\[
\frac{2}{|\mathcal{Y}||\mathcal{G}|} \sum_{y \in \mathcal{Y}} \sum_{g \in \mathcal{G}} 
\delta_{\mathrm{TV}}\bigl(P^{\hat f_\theta(\mathbf{X}) \mid y,g}_{d_T},\, P^{\hat f_\theta(\mathbf{X}) \mid y,g}\bigr) \leq \frac{2}{|\mathcal{Y}||\mathcal{G}|} \sum_{y \in \mathcal{Y}} \sum_{g \in \mathcal{G}} 
\delta_{\mathrm{TV}}\bigl(P^{\mathbf{X} \mid y,g}_{d_T},\, P^{\mathbf{X} \mid y,g}\bigr)
\]

For \textbf{Term (3)}, we have 
\[
\begin{aligned}
\textbf{Term (3)}
&=\frac{2}{|\mathcal{Y}||\mathcal{G}|}\sum_{d_S \in \mathcal{D}_S}\sum_{y\in\mathcal{Y}}\sum_{g\in\mathcal{G}}\frac{p(d_S,y,g)}{p(y,g)}
\delta_{\rm TV}\bigl(P^{\hat f_\theta(\mathbf{X}) \mid y,g}_{d_S},P^{\hat f_\theta(\mathbf{X}) \mid y,g}\bigr)\\&\le \frac{2}{|\mathcal{Y}||\mathcal{G}|\displaystyle \min_{\substack{y, g}}p(y,g)}
\sum_{d_S \in \mathcal{D}_S}\sum_{y\in\mathcal{Y}}\sum_{g\in\mathcal{G}}p(d_S,y,g)
\delta_{\rm TV}\bigl(P^{\hat f_\theta(\mathbf{X}) \mid y,g}_{d_S},P^{\hat f_\theta(\mathbf{X}) \mid y,g}\bigr).
\end{aligned}
\]

Applying Lemma 4, with the identification 
\(\mathbf X\!\mapsto\!\mathbf D_S\), 
\(\mathbf Y\!\mapsto\!\hat f_\theta(\mathbf{X})\), 
\(\mathbf D\!\mapsto\!\mathbf Y\), 
and \(\mathbf G\!\mapsto\!\mathbf G\),  
one has
\[
\sum_{d_S \in \mathcal{D}_S}\sum_{y\in\mathcal{Y}}\sum_{g\in\mathcal{G}}p(d_S,y,g)
\delta_{\rm TV}\bigl(P^{\hat f_\theta(\mathbf{X}) \mid y,g}_{d_S},\,P^{\hat f_\theta(\mathbf{X}) \mid y,g}\bigr)
\;\le\;
\sqrt{\frac{I\bigl(\mathbf D_S;\,\hat f_\theta(\mathbf{X})\mid\mathbf Y,\mathbf G\bigr)}{2}}.
\]

Combining the two displays shows that \textbf{Term (3)} is upper bounded by
\[
\textbf{Term (3)} \leq \frac{\sqrt{2I\bigl(\mathbf D_S;\,\hat f_\theta(\mathbf{X})\mid\mathbf Y,\mathbf G\bigr)}}{|\mathcal{Y}||\mathcal{G}|\displaystyle \min_{\substack{y, g}}p(y,g)}.
\]

Combining the upper bounds of \textbf{Term (1)}, \textbf{(2)}, and \textbf{(3)} with Eq. (\textcolor{red}{$\dagger$}), we have:
\[
 \Delta_{d_T}^{\mathrm{EOD}}(\hat{f}_\theta) \leq \frac{\sqrt{2I\bigl(\mathbf G;\,\hat f_\theta(\mathbf{X})\mid\mathbf Y,\mathbf D_S\bigr)}}{|\mathcal{Y}||\mathcal{G}|\displaystyle \min_{\substack{y, g}}p(y,g)} + \frac{2}{|\mathcal{Y}||\mathcal{G}|} \sum_{y \in \mathcal{Y}} \sum_{g \in \mathcal{G}} 
\delta_{\mathrm{TV}}\bigl(P^{\mathbf{X} \mid y,g}_{d_T},\, P^{\mathbf{X} \mid y,g}\bigr) + \frac{\sqrt{2I\bigl(\mathbf D_S;\,\hat f_\theta(\mathbf{X})\mid\mathbf Y,\mathbf G\bigr)}}{|\mathcal{Y}||\mathcal{G}|\displaystyle \min_{\substack{y, g}}p(y,g)}.
\]
\qed \\

\noindent\textbf{Proof of Theorem 3.} We begin by defining the expected risk under the cross-entropy loss and the mutual information between the model prediction and the true label:
\[
\mathcal{R}(\hat{f}_\theta) = \mathbb{E}_{(\mathbf{X},\mathbf{Y})} \left[ -\log \hat{f}_\theta(\mathbf{X})_\mathbf{Y} \right], \quad \text{and} \quad  I(\hat{f}_\theta(\mathbf{X}); \mathbf{Y}) = H(\mathbf{Y}) - H(\mathbf{Y} | \hat{f}_\theta(\mathbf{X})),
\]

\noindent where \(\hat{f}_\theta(\mathbf{X})_\mathbf{Y}\) denotes the predicted probability assigned to the true label \(\mathbf{Y}\) and $H(\mathbf{Y} | \hat{f}_\theta(\mathbf{X})) = \mathbb{E}_{(\mathbf{X},\mathbf{Y})}[-\log P(\mathbf{Y}|\hat{f}_\theta(\mathbf{X}))]$. 

Subtracting $H(\mathbf{Y}\mid \hat{f}_\theta(\mathbf{X}))$ from the risk yields

\[
\begin{aligned}
\mathcal{R}(\hat{f}_\theta) - H(\mathbf{Y} | \hat{f}_\theta(\mathbf{X})) &= \mathbb{E}_{(\mathbf{X},\mathbf{Y})}\left[-\log \hat{f}_\theta(\mathbf{X})_\mathbf{Y} +\log P(\mathbf{Y}|\hat{f}_\theta(\mathbf{X}))\right]
\\&= \mathbb{E}_{(\mathbf{X},\mathbf{Y})}\left[\log\frac{P(\mathbf{Y}|\hat{f}_\theta(\mathbf{X}))}{ \hat{f}_\theta(\mathbf{X})_\mathbf{Y}}\right]
\\&\stackrel{(1)}{=} 
\mathbb{E}_{(\hat{f}_\theta(\mathbf{X}),\mathbf{Y})}\left[\log\frac{P(\mathbf{Y}|\hat{f}_\theta(\mathbf{X}))}{ \hat{f}_\theta(\mathbf{X})_\mathbf{Y}}\right]
\\&\stackrel{(2)}{=} \mathbb{E}_{\hat{f}_\theta(\mathbf{X})}\left[\mathbb{E}_{\mathbf{Y}|\hat{f}_\theta(\mathbf{X})}\left[\log\frac{P(\mathbf{Y}|\hat{f}_\theta(\mathbf{X}))}{ \hat{f}_\theta(\mathbf{X})_\mathbf{Y}}\right]\right]
\\&\stackrel{(3)}{=} \mathbb{E}_{\hat{f}_\theta(\mathbf{X})}\left[D_{KL}(P||Q)\right]
\\&\stackrel{(4)}{\geq} 0
\end{aligned}
\]
Here, $\stackrel{(1)}{=}$ is due to the Law of the Unconscious Statistician (LOTUS) by the measurable map $(\mathbf{X},\mathbf{Y})\!\mapsto\!(\hat{f}_\theta(\mathbf{X}),\mathbf{Y})$. $\stackrel{(2)}{=}$ is due to the Tower property $\mathbb{E}_{(\mathbf{X},\mathbf{Y})}[\cdot]=\mathbb{E}_{\mathbf{X}}\!\bigl[\mathbb{E}_{\mathbf{Y}\mid \mathbf{X}}[\cdot]\bigr]$.  $\stackrel{(3)}{=}$ is the definition of KL divergence:  $\sum_y P(y)\log\frac{P(y)}{Q(y)}$, where $P(y) = P(\mathbf{Y}=y|\hat{f}_\theta(\mathbf{X}))$ and $Q(y) = \hat{f}_\theta(\mathbf{X})_y$. $\stackrel{(4)}{\geq}$ is the Non-negativity of KL.\\

Thus, we have $\mathcal{R}(\hat{f}_\theta)$ lower bounded by $H(\mathbf{Y} | \hat{f}_\theta(\mathbf{X}))$:
\[
\mathcal{R}(\hat{f}_\theta) \geq  H(\mathbf{Y} | \hat{f}_\theta(\mathbf{X})).
\]

Equality holds \emph{iff}
\[
P(\mathbf{Y}\mid\hat{f}_\theta(\mathbf{X}))
=\hat{f}_\theta(\mathbf{X})_\mathbf{Y}
=\!P(\mathbf{Y}\mid\mathbf{X})\tag{BO}
\]
Condition~(BO) is precisely \textbf{Bayes-optimality}: the predictor
outputs the true conditional label distribution ($D_{KL}(P||Q)=0$).

Observing that $H(\mathbf{Y} | \hat{f}_\theta(\mathbf{X})) =  H(\mathbf{Y}) - I(\hat{f}_\theta(\mathbf{X}); \mathbf{Y})$, we have:

\[
I(\hat{f}_\theta(\mathbf{X}); \mathbf{Y}) \geq H(\mathbf{Y}) - \mathcal{R}(\hat{f}_\theta).
\]

In the domain generalization setting, we only have access to data from the source domains. We therefore condition on a fixed source domain \(d_S\) and then average over
\(d_S\in\mathcal{D}_S\):
\[
\sum_{d_S \in \mathcal{D}_S}p(d_S)I(\hat{f}_\theta(\mathbf{X}); \mathbf{Y}|\mathbf{D_S}=d_S) \geq \sum_{d_S \in \mathcal{D}_S}p(d_S)H(\mathbf{Y}|\mathbf{D}_S=d_S) - \sum_{d_S \in \mathcal{D}_S}p(d_S)\mathcal{R}_{d_S}(\hat{f}_\theta).
\]

This simplifies to:

\[
I(\hat{f}_\theta(\mathbf{X}); \mathbf{Y}|\mathbf{D_S}) \geq H(\mathbf{Y}|\mathbf{D}_S) - \mathcal{R}_{\mathcal{D}_S}(\hat{f}_\theta).
\]

Combining this lower bound with the data-processing inequality $I(\hat{f}_\theta(\mathbf{X});\mathbf{Y}\mid\mathbf{D_S})
\;\le\;
I(\mathbf{X};\mathbf{Y}\mid\mathbf{D_S})$, we obtain the information sandwich, for every
\(\theta\in\Theta\),

\[
H(\mathbf{Y}\mid\mathbf{D_S})-\mathcal{R}_{\mathcal{D_S}}(\hat{f}_\theta)
\stackrel{(1)}{\leq}
I(\hat{f}_\theta(\mathbf{X});\mathbf{Y}\mid\mathbf{D_S})
\stackrel{(2)}{\leq}
I(\mathbf{X};\mathbf{Y}\mid\mathbf{D_S}).
\]

\noindent\textit{Risk minimiser $\Longrightarrow$ MI maximiser.} Let
\(
\theta^\star=
\arg\min_\theta\mathcal{R}_{\mathcal{D_S}}(\hat{f}_\theta)
\)
and write
\(
\mathcal{R}_{\min}=
\mathcal{R}_{\mathcal{D_S}}(\hat{f}_{\theta^\star}).
\)
Because \(\theta^\star\) attains the global minimum, Bayes-optimality (BO) holds for \(\theta^\star\), which makes $\stackrel{(1)}{\leq}$ tight, giving
\[
I(\hat{f}_{\theta^\star}(\mathbf{X});\mathbf{Y}\mid\mathbf{D_S})
=
H(\mathbf{Y}\mid\mathbf{D_S})-\mathcal{R}_{\min}.
\]
Moreover, (BO) implies
\(
P(\mathbf{Y}\mid\hat{f}_\theta(\mathbf{X}))
=\!P(\mathbf{Y}\mid\mathbf{X})
\),
so data-processing is also tight:
\[
I(\hat{f}_{\theta^\star}(\mathbf{X});\mathbf{Y}\mid\mathbf{D_S})
=
I(\mathbf{X};\mathbf{Y}\mid\mathbf{D_S}),
\]
the largest value allowed by $\stackrel{(2)}{\leq}$.  
Hence every risk minimiser is a mutual-information maximiser.

\noindent\textit{MI maximiser $\Longrightarrow$ risk minimiser.} Conversely, let
\(
\bar\theta=
\arg\max_\theta
I(\hat{f}_\theta(\mathbf{X});\mathbf{Y}\mid\mathbf{D_S}).
\)
Equality on the $\stackrel{(2)}{\leq}$ must then hold, so
\(
I(\hat{f}_{\bar\theta}(\mathbf{X});\mathbf{Y}\mid\mathbf{D_S})
      =I(\mathbf{X};\mathbf{Y}\mid\mathbf{D_S})
\).
This forces Bayes-optimality
\(
P(\mathbf{Y}\mid\hat{f}_\theta(\mathbf{X}))
=\!P(\mathbf{Y}\mid\mathbf{X})
\)
which makes $D_{KL}(P||Q)=0$, implying
\(
\mathcal{R}_{\mathcal{D_S}}(\hat{f}_{\bar\theta})
      =\mathcal{R}_{\min}.
\)
Thus, every MI maximiser is also a risk minimiser.\\

Combining the two inclusions gives
\[
\arg\min_{\theta\in\Theta}\mathcal{R}_{\mathcal{D_S}}(\hat{f}_\theta)
\;=\;
\arg\max_{\theta\in\Theta}
I(\hat{f}_\theta(\mathbf{X});\mathbf{Y}\mid\mathbf{D_S}).
\]

\qed \\

\noindent\textbf{Proof of Theorem 4.} For the first inequality, consider two chain-rule decompositions of the mutual information $I(\hat{f}_\theta(\mathbf{X});\mathbf{D}_S, \mathbf{G}| \mathbf{Y})$:
\[
I(\hat{f}_\theta(\mathbf{X});\mathbf{D}_S, \mathbf{G}| \mathbf{Y}) =
        I(\hat{f}_\theta(\mathbf{X});\mathbf{D}_S| \mathbf{Y}) +
        I(\hat{f}_\theta(\mathbf{X});\mathbf{G}| \mathbf{Y},\mathbf{D}_S)=I(\hat{f}_\theta(\mathbf{X});\mathbf{G}|  \mathbf{Y}) + I(\hat{f}_\theta(\mathbf{X});\mathbf{D}_S| \mathbf{Y},\mathbf{G}).
\]
Equating the two decompositions, rearranging terms, and using the symmetry property of MI, we obtain:
\[
I(\mathbf{D}_S;\hat{f}_\theta(\mathbf{X})\mid \mathbf{Y},\mathbf{G})
  = I(\mathbf{D}_S;\hat{f}_\theta(\mathbf{X})\mid \mathbf{Y})
+ I(\mathbf{G};\, \hat{f}_\theta(\mathbf{X}) \mid \mathbf{Y}, \mathbf{D}_S)
  - I(\mathbf{G};\, \hat{f}_\theta(\mathbf{X}) \mid \mathbf{Y}). 
\]
Since MIs are non-negative, it follows that:
\[
I(\mathbf{D}_S;\hat{f}_\theta(\mathbf{X})\mid \mathbf{Y},\mathbf{G})
  \leq I(\mathbf{D}_S;\hat{f}_\theta(\mathbf{X})\mid \mathbf{Y})
  + I(\mathbf{G};\, \hat{f}_\theta(\mathbf{X}) \mid \mathbf{Y}, \mathbf{D}_S). 
\]

For the second inequality, considering the two chain-rule decompositions of $I(\hat f_\theta(\mathbf X);\mathbf Y,\mathbf D_S)$:
\[
I\bigl(\hat f_\theta(\mathbf X);\mathbf Y,\mathbf D_S\bigr)
=I\bigl(\hat f_\theta(\mathbf X);\mathbf D_S\bigr)
+I\bigl(\hat f_\theta(\mathbf X);\mathbf Y\mid \mathbf D_S\bigr)
=I\bigl(\hat f_\theta(\mathbf X);\mathbf Y\bigr)
+I\bigl(\hat f_\theta(\mathbf X);\mathbf D_S\mid \mathbf Y\bigr).
\]
Equating the two decompositions, rearranging terms, and using the symmetry property of MI, we obtain:
\[
I\bigl(\hat f_\theta(\mathbf X);\mathbf Y\bigr)=I\bigl(\hat f_\theta(\mathbf X);\mathbf D_S\bigr)+
I\bigl(\hat f_\theta(\mathbf X);\mathbf Y\mid \mathbf D_S\bigr)-
I\bigl(\mathbf D_S;\hat f_\theta(\mathbf X)\mid \mathbf Y\bigr).
\]

Since MIs are non-negative, it follows that:
\[
I\bigl(\hat f_\theta(\mathbf X);\mathbf Y\bigr)\geq 
I\bigl(\hat f_\theta(\mathbf X);\mathbf Y\mid \mathbf D_S\bigr)-
I\bigl(\mathbf D_S;\hat f_\theta(\mathbf X)\mid \mathbf Y\bigr).
\]

For the third inequality, considering the chain-rule decomposition of $I(\mathbf D_S,\mathbf G;\hat f_\theta(\mathbf X)|\mathbf Y)$:
\[
I(\mathbf D_S,\mathbf G;\hat f_\theta(\mathbf X)|\mathbf Y)
=
I(\mathbf D_S;\hat f_\theta(\mathbf X)|\mathbf Y)+
I(\mathbf G;\hat f_\theta(\mathbf X)|\mathbf Y,\mathbf D_S)=
I(\mathbf G;\hat f_\theta(\mathbf X)|\mathbf Y)+
I(\mathbf D_S;\hat f_\theta(\mathbf X)|\mathbf Y,\mathbf G).
\]
Equating the two decompositions and rearranging, we have:
\[
I(\mathbf G;\hat f_\theta(\mathbf X)\mid \mathbf Y)=I(\mathbf D_S;\hat f_\theta(\mathbf X)\mid \mathbf Y)+
I(\mathbf G;\hat f_\theta(\mathbf X)\mid \mathbf Y,\mathbf D_S)-
I(\mathbf D_S;\hat f_\theta(\mathbf X)\mid \mathbf Y,\mathbf G). 
\]
Since MIs are non-negative, it follows that:
\[
I(\mathbf G;\hat f_\theta(\mathbf X)\mid \mathbf Y)\leq I(\mathbf D_S;\hat f_\theta(\mathbf X)\mid \mathbf Y)+
I(\mathbf G;\hat f_\theta(\mathbf X)\mid \mathbf Y,\mathbf D_S). 
\]

\qed \\

\noindent\textbf{\Large  \textcolor{red}{C}. Proofs for the Lemmas}\\

\noindent\textbf{Proof of Lemma 1.}  We provide the proof for the case where $\mathbf{X}$ is a continuous random variable; a similar argument holds for the discrete case by replacing integrals with summations accordingly. Consider two domains \( d_j \) and \( d_i \) with corresponding probability density functions \( p_{d_j}(x) \) and \( p_{d_i}(x) \) over \( \mathbf{X} \). Define \( p_{j \rightarrow i}(x) = p_{d_j}(x) - p_{d_i}(x) \) and \( p_{i \rightarrow j}(x) =p_{d_i}(x) - p_{d_j}(x)= -p_{j \rightarrow i}(x) \). Let \( \mathcal{X}^+ := \{ x \in \mathcal{X} : p_{j \to i}(x) > 0 \} \) and \( \mathcal{X}^- := \{ x \in \mathcal{X} : p_{j \to i}(x) \leq 0 \} \). Then we can write:
\[
\mathbb{E}_{d_j}[f(\mathbf{X})] - \mathbb{E}_{d_i}[f(\mathbf{X})]= \int_{\mathcal{X}} f(x)\, (p_{d_j}(x) - p_{d_i}(x)) \, dx  =\int_{\mathcal{X}^+} f(x)\, p_{j \to i}(x) \, dx + \int_{\mathcal{X}^-} f(x)\, p_{j \to i}(x) \, dx.
\]

Since \( 0 \leq f(x) \leq C \) for all \( x \in \mathcal{X} \), and \( p_{j \rightarrow i}(x) \leq 0 \) on \( \mathcal{X}^- \), making the second integral non-positive, we have:
\[
\mathbb{E}_{d_j}[f(\mathbf{X})] - \mathbb{E}_{d_i}[f(\mathbf{X})] \leq C \int_{\mathcal{X}^+} p_{j \to i}(x)\, dx.
\]

To proceed, observe that:
\[
\int_{\mathcal{X}} p_{j\rightarrow i}(x) dx = \int_{\mathcal{X}} p_{d_j}(x) dx - \int_{\mathcal{X}}p_{d_i}(x) \, dx = 1-1= 0,
\]
and also:
\[
\int_{\mathcal{X}} p_{j \to i}(x)\, dx 
= \int_{\mathcal{X}^+} p_{j \to i}(x)\, dx + \int_{\mathcal{X}^-} p_{j \to i}(x)\, dx 
= \int_{\mathcal{X}^+} p_{j \to i}(x)\, dx - \int_{\mathcal{X}^-} p_{i \to j}(x)\, dx.
\]

Hence,
\[
\int_{\mathcal{X}^+} p_{j \to i}(x)\, dx = \int_{\mathcal{X}^-} p_{i \to j}(x)\, dx.
\]

Therefore, we have:
\[
\mathbb{E}_{d_j}[f(\mathbf{X})] - \mathbb{E}_{d_i}[f(\mathbf{X})] \leq C \int_{\mathcal{X}^+} p_{j \to i}(x)\, dx
 = \frac{C}{2} \int_{\mathcal{X}^+} p_{j \to i}(x)\, dx + \frac{C}{2} \int_{\mathcal{X}^-} p_{i \to j}(x)\, dx.
\]

Noting that:
\[
\int_{\mathcal{X}} |p_{d_j}(x) - p_{d_i}(x)|\, dx = \int_{\mathcal{X}^+} |p_{j \to i}(x)|\, dx + \int_{\mathcal{X}^-} |p_{j \to i}(x)|\, dx = \int_{\mathcal{X}^+} p_{j \to i}(x)\, dx + \int_{\mathcal{X}^-} p_{i \to j}(x)\, dx,
\]
\noindent we conclude:
\[
\mathbb{E}_{d_j}[f(\mathbf{X})] - \mathbb{E}_{d_i}[f(\mathbf{X})] \leq \frac{C}{2} \int_{\mathcal{X}} |p_{d_j}(x) - p_{d_i}(x)|\, dx = C \cdot \delta_{\mathrm{TV}}(P_{d_j}^{\mathbf{X}}, P_{d_i}^{\mathbf{X}}),
\]

\noindent where \( \delta_{\mathrm{TV}}(P_{d_j}^{\mathbf{X}}, P_{d_i}^{\mathbf{X}}) = \frac{1}{2} \int_{\mathcal{X}} \left| p_{d_j}(x) - p_{d_i}(x) \right| \, dx \) is the total variation (TV) distance between $P_{d_j}^{\mathbf{X}}$ and $P_{d_i}^{\mathbf{X}}$.\\

\hfill $\qed$ \\

\noindent \textbf{Proof of Lemma 2.}  We provide the proof for the case where $\mathbf{Y}$ is a continuous random variable; a similar argument holds for the discrete case by replacing integrals with summations accordingly. By applying Pinsker’s inequality and Jensen’s inequality of the concave function \( \sqrt{\cdot} \), we have:
\[
\sum_{x \in \mathcal{X}} p(x)\, \delta_{\mathrm{TV}}\!\Bigl(P_{\mathbf{Y}|x},\, P_\mathbf{Y}\Bigr) \leq 
\sum_{x \in \mathcal{X}}  p(x)\, \sqrt{\frac{1}{2} D_{\mathrm{KL}}\!\Bigl(P_{\mathbf{Y}|x},\, P_\mathbf{Y}\Bigr)} \leq \sqrt{ \frac{1}{2} \sum_{x \in \mathcal{X}}  p(x)\, D_{\mathrm{KL}}\!\Bigl(P_{\mathbf{Y}|x},\, P_\mathbf{Y}\Bigr) }.
\]

Since \( \mathbf{Y} \) is continuous, the KL divergence between \( P_{\mathbf{Y}|x} \) and \( P_\mathbf{Y} \) is defined via integrals. Using Fubini’s theorem to interchange the summation and the integral, we have:
\[
\sum_{x \in \mathcal{X}} p(x)\, D_{\mathrm{KL}}\!\Bigl(P_{\mathbf{Y}|x},\, P_\mathbf{Y}\Bigr)=\sum_{x \in \mathcal{X}}  p(x) \int_{\mathcal{Y}} p(y \mid x) \log \frac{p(y \mid x)}{p(y)} \, dy = 
\int_{\mathcal{Y}} \sum_{x \in \mathcal{X}} p(x)  p(y \mid x) \log \frac{p(y \mid x)}{p(y)} \, dy.
\]

Noting that \( p(x, y) = p(x)\, p(y \mid x) \), we rewrite the expression as:
\[
\int_{\mathcal{Y}} \sum_{x \in \mathcal{X}} p(x, y) \log \frac{p(y \mid x)}{p(y)} \, dy 
= 
\int_{\mathcal{Y}} \sum_{x \in \mathcal{X}} p(x, y) \log \frac{p(x, y)}{p(x) p(y)} \, dy 
= 
D_{\mathrm{KL}}\!\bigl(P_{\mathbf{X}, \mathbf{Y}} \,\|\, P_\mathbf{X} P_\mathbf{Y}\bigr) = I(\mathbf{X}; \mathbf{Y}).
\]

Thus, we conclude:
\[
\sum_{x \in \mathcal{X}} p(x)\, \delta_{\mathrm{TV}}\!\Bigl(P_{\mathbf{Y}|x},\, P_\mathbf{Y}\Bigr) 
\leq 
\sqrt{ \frac{I(\mathbf{X}; \mathbf{Y})}{2} }.
\]
\hfill $\qed$ \\

\noindent \textbf{Proof of Lemma 3.} We provide the proof for continuous random variables; a similar argument holds for the discrete case by replacing integrals with summations accordingly.  Let $\mathcal X\subseteq\mathbb R^n$ be the support of $\mathbf X$, and write $p_{d_i}(x)$, $p'_{d_i}(x)$, $p_{d_j}(x)$, $p'_{d_j}(x)$ for the corresponding probability density functions.  Then
\[
\begin{aligned}
\delta_{\mathrm{TV}}\bigl(P_{d_j},\,P'_{d_j}\bigr)\;-\;\delta_{\mathrm{TV}}\bigl(P_{d_i},\,P'_{d_i}\bigr) &= \frac12\int_{\mathcal X}\bigl|p_{d_j}(x)-p'_{d_j}(x)\bigr| \;-\;\bigl|p_{d_i}(x)-p'_{d_i}(x)\bigr|\,dx \\
&\stackrel{(1)}{\le} \frac12\int_{\mathcal X}
     \bigl|p_{d_j}(x)-p'_{d_j}(x) - (p_{d_i}(x)-p'_{d_i}(x))\bigr|\,dx \\
&= \frac12\int_{\mathcal X}
     \bigl|\,p_{d_j}(x)-p_{d_i}(x)\;-\;(p'_{d_j}(x)-p'_{d_i}(x))\bigr|\,dx \\
&\stackrel{(2)}{\le} \frac12\int_{\mathcal X}
     \bigl|p_{d_j}(x)-p_{d_i}(x)\bigr| + \bigl|p'_{d_j}(x)-p'_{d_i}(x)\bigr|\,dx \\
&= \delta_{\mathrm{TV}}\bigl(P_{d_j},\,P_{d_i}\bigr)
  \;+\;
  \delta_{\mathrm{TV}}\bigl(P'_{d_j},\,P'_{d_i}\bigr).
\end{aligned}
\]

Here, $\stackrel{(1)}{\leq}$ and $\stackrel{(2)}{\leq}$ are because of the triangle inequality $|a|-|b| \leq |a-b| \leq |a|+|b|$. 

\hfill $\qed$ \\

\noindent\textbf{Proof of Lemma 4.}  
We first rewrite Lemma 2 conditioned on \( (d,\,g) \), by making the following substitutions:
\[
p(x) \to p(x \mid d,\,g), \quad
P_{\mathbf{Y} \mid x} \to P_{\mathbf{Y} \mid x,\,d,\, g}, \quad P_\mathbf{Y} \to P_{\mathbf{Y} \mid d,\, g}, \quad
I(\mathbf{X}; \mathbf{Y}) \to I(\mathbf{X}; \mathbf{Y} \mid d,\, g).
\]

This yields:
\[
\sum_{x\in\mathcal{X}}
p(x\mid d,g)\;
\delta_{\mathrm{TV}}\!\bigl(P_{\mathbf{Y}\mid x,d,g},\,P_{\mathbf{Y}\mid d,g}\bigr)
\;\le\;
\sqrt{\frac{I\bigl(\mathbf{X};\mathbf{Y}\mid d,g\bigr)}{2}}.
\]

Multiplying both sides by \( p(d, g) \) and summing over \( d, g \) gives
\[
\sum_{d \in \mathcal{D}}\sum_{g \in \mathcal{G}} p(d, g)\,
\sum_{x \in \mathcal{X}} p(x \mid d, g)\,
\delta_{\mathrm{TV}}\bigl(P_{\mathbf{Y} \mid x, d, g},\, P_{\mathbf{Y} \mid d, g)}\bigr)
\;\le\;
\sum_{d \in \mathcal{D}}\sum_{g \in \mathcal{G}} p(d, g)\,
\sqrt{\frac{I(\mathbf{X}; \mathbf{Y} \mid d, g)}{2}}.
\]

Noting that $p(d, g)p(x \mid d, g)= p(x, d, g)$, for the left-hand side, we have:
\[
\sum_{d \in \mathcal{D}}\sum_{g \in \mathcal{G}} p(d, g)\,
\sum_{x \in \mathcal{X}} p(x \mid d, g)\,
\delta_{\mathrm{TV}}\bigl(P_{\mathbf{Y} \mid x, d, g},\, P_{\mathbf{Y} \mid d, g)}\bigr)
=
\sum_{x \in \mathcal{X}}\sum_{d \in \mathcal{D}}\sum_{g \in \mathcal{G}}  p(x, d, g)\,
\delta_{\mathrm{TV}}\bigl(P_{\mathbf{Y} \mid x, d, g},\, P_{\mathbf{Y} \mid d, g)}\bigr).
\]

For the right-hand side, since $u\mapsto\sqrt{u}$ is concave, Jensen’s inequality yields
\[
\sum_{d \in \mathcal{D}}\sum_{g \in \mathcal{G}} p(d, g)\,
\sqrt{\frac{I(\mathbf{X}; \mathbf{Y} \mid d, g)}{2}}
\;\le\;
\sqrt{\frac12\,
\sum_{d \in \mathcal{D}}\sum_{g \in \mathcal{G}}p(d,g)\,I(\mathbf{X};\mathbf{Y}\mid d,g)}
\;=\;
\sqrt{\frac{I(\mathbf{X};\mathbf{Y}\mid\mathbf{D},\mathbf{G})}{2}}.
\]

Combining the above, we have:
\[
\sum_{x \in \mathcal{X}}\sum_{d \in \mathcal{D}}\sum_{g \in \mathcal{G}} p(x, d, g)\,
\delta_{\mathrm{TV}}\bigl(P_{\mathbf{Y} \mid x, d, g},\, P_{\mathbf{Y} \mid d, g)}\bigr)
\;\le\;
\sqrt{\frac{I(\mathbf{X};\mathbf{Y}\mid\mathbf{D},\mathbf{G})}{2}}.
\]
\qed \\

\noindent\textbf{\Large  \textcolor{red}{D}. Comparisons with the Theoretical Bounds}\\

\textbf{\large Previous bounds in domain generalization}

\begin{itemize}
    \item Bounds in \cite{albuquerque2019generalizing}:
\end{itemize}
\[
\epsilon_{D^T}^{\text{Acc}}\left(\hat{f}\right) \leq 
\sum_{i=1}^{N} \pi_i \epsilon_{D^S_i}^{\text{Acc}}\left(\hat{f}\right) 
+ \max_{j,k \in [N]} \mathcal{D}_{\mathcal{H}}\left(P_{D^S_j}^X \parallel P_{D^S_k}^X\right)
+ \mathcal{D}_{\mathcal{H}}\left(P_{D^S_*}^X \parallel P_{D^T}^X\right)
+ \min_{D \in \{D^S_*, D^T\}} \mathbb{E}_D\left[ \left| f_{D^S_*}(X) - f_{D^T}(X) \right| \right]
\]

where
\[
\mathcal{D}_{\mathcal{H}}\left(P_{D^S}^X \parallel P_{D^T}^X\right) = \sup_{\hat{f}} \left| P_{D^S}\left(\hat{f}(X)=1\right) - P_{D^T}\left(\hat{f}(X)=1\right) \right|
\]
is the $\mathcal{H}$ divergence,
\[
P_{D^S_*}^X = \arg\min_{\pi} \mathcal{D}_{\mathcal{H}}\left(\sum_{i=1}^{N} \pi_i P_{D^S_i}^X \parallel P_{D^T}^X\right)
\]
is the mixture of source domains closest to the target domain under $\mathcal{H}$-divergence.

In this bound, the target domain classification error is upper bounded by four terms: (1) a convex combination of errors in the source domains, (2) the $\mathcal{H}$-divergence between source domains, (3) the $\mathcal{H}$-divergence between the target domain and its closest source domain mixture, and (4) the discrepancy between labeling functions in the source mixture and target domain. Since the target domain is unknown in domain generalization, terms involving $D^T$ are uncontrollable. Algorithmic designs therefore typically focus on minimizing source domain errors and reducing the $\mathcal{H}$-divergence between source domains.

\begin{itemize}
    \item Bounds in \cite{phung2021learning}:
\end{itemize}

\[
\epsilon_{D^T}^{\text{Acc}}\left(\hat{f}\right) \leq 
\sum_{i=1}^{N} \pi_i \epsilon_{D^S_i}^{\text{Acc}}\left(\hat{f}\right)
+ C \max_{i \in [N]} \mathbb{E}_{D^S_i} \left[ \left\| \left[ \left| f_{D^T}(X)_y - f_{D^S_i}(X)_y \right| \right]_{y=1}^{|\mathcal{Y}|} \right\|_1 \right]
\]
\[
+ \sum_{i=1}^{N} \sum_{j=1}^{N} \frac{C \sqrt{2\pi_j}}{N} d_{1/2}\left(P_{D^T}^Z, P_{D^S_i}^Z\right)
+ \sum_{i=1}^{N} \sum_{j=1}^{N} \frac{C \sqrt{2\pi_j}}{N} d_{1/2}\left(P_{D^S_i}^Z, P_{D^S_j}^Z\right)
\]

where
\[
d_{1/2}\left(P_{D^S_i}^X, P_{D^S_j}^X\right) = \sqrt{\mathcal{D}_{1/2}\left(P_{D^S_i}^X \parallel P_{D^S_j}^X\right)}
\]
is Hellinger distance defined based on

\[
\mathcal{D}_{1/2}\left(P_{D^S_i}^X \parallel P_{D^S_j}^X\right) = 2 \int_X \left(\sqrt{P_{D^S_i}^X} - \sqrt{P_{D^S_j}^X}\right)^2 dX.
\]

As in the previous case, terms involving $D^T$ are beyond control in domain generalization. Thus, algorithmic efforts typically focus on minimizing the convex combination of source domain errors and reducing the Hellinger distances between source domains.

\begin{itemize}
    \item Bounds in \cite{pham2023fairness}:
\end{itemize}

\[
\epsilon_{D^T}^{\text{Acc}}\left(\hat{f}\right) \leq 
\frac{1}{N} \sum_{i=1}^{N} \epsilon_{D^S_i}^{\text{Acc}}\left(\hat{f}\right)
+ \sqrt{2}C \min_{i \in [N]} d_{JS}\left(P_{D^T}^{X,Y}, P_{D^S_i}^{X,Y}\right)
+ \sqrt{2}C \max_{i,j \in [N]} d_{JS}\left(P_{D^S_i}^{Z,Y}, P_{D^S_j}^{Z,Y}\right)
\]

where $d_{JS}(\cdot,\cdot)$ denotes the Jensen-Shannon distance.

Again, because $D^T$ is unavailable during training, algorithmic focus shifts to minimizing the average source domain errors and reducing the JS distances between source domains.\\

\noindent\textbf{Summary:} As discussed above, prior bounds in domain generalization rely heavily on distribution matching using metrics such as $\mathcal{H}$-divergence, Hellinger distance, or JS distance. A key limitation of these approaches is poor scalability: as the number of classes and source domains increases, the number of distributions to align grows as $|\mathcal{Y}|\times|\mathcal{D}_S|$.  In contrast, we are the first to derive a domain generalization bound based on mutual information, which avoids extensive distribution alignment. Our upper bound better supports algorithm design for complex DG settings with multi-class tasks and many source domains, enabling methods that scale well to real-world applications. \\

\noindent\textbf{\large Previous fairness bounds in domain generalization}\\

\cite{pham2023fairness} is the first work to derive fairness upper bounds in the domain generalization setting (see Theorem 3 in their paper). The result is presented below:

\textbf{Theorem 3 (Upper bound: fairness)} Consider a special case where the unfairness measure is defined as the distance between means of two distributions:  

\[
\epsilon_{D}^{EO}(\hat{f}) = \sum_{y \in \{0,1\}} \left\| \mathbb{E}_{D}\left[\hat{f}(X)_1 \mid Y = y, A = 0\right] - \mathbb{E}_{D}\left[\hat{f}(X)_1 \mid Y = y, A = 1\right] \right\|,
\]

then the unfairness at any unseen target domain $D^T$ is upper bounded:

\[
\epsilon_{D^T}^{EO}(\hat{f}) \leq
\frac{1}{N} \sum_{i=1}^{N} \epsilon_{D^S_i}^{EO}(\hat{f})
+ \sqrt{2} \min_{i \in [N]} \sum_{y \in \{0,1\}} \sum_{a \in \{0,1\}} d_{JS}\left(P_{D^T}^{X|Y=y,A=a}, P_{D^S_i}^{X|Y=y,A=a}\right)
\]
\[
+ \sqrt{2} \max_{i,j \in [N]} \sum_{y \in \{0,1\}} \sum_{a \in \{0,1\}} d_{JS}\left(P_{D^S_i}^{Z|Y=y,A=a}, P_{D^S_j}^{Z|Y=y,A=a}\right)
\]

where $d_{JS}(\cdot,\cdot)$ denotes the Jensen-Shannon distance.

As in domain generalization, $D^T$ is unknown, so the second term involving $D^T$ is uncontrollable. Consequently, the main message from this bound is to minimize fairness violations (in this case, Equal Opportunity) within and across the source domains by JS distance. \\

\noindent\textbf{Summary:} This fairness bound, like the domain generalization bounds in their work, relies on distribution matching. As discussed earlier, it suffers from scalability issues in multi-class tasks with multi-group sensitive attributes, since the number of distributions to align grows as $|\mathcal{Y}|\times|\mathcal{G}|$. As a result, their fairness bound is limited to binary classification with binary-group sensitive attributes. Moreover, it is based on only matching distribution expectations, which is insufficient to capture fairness metrics such as EO and EOD that depend on aligning entire conditional distributions. In contrast, we propose a fairness bound based on mutual information, which scales naturally to multi-class tasks with multi-group sensitive attributes and directly aligns with fairness definitions like EO and EOD, as these can be expressed in terms of enforcing conditional statistical independence. \\

\noindent\textbf{\Large  \textcolor{red}{E}. Calculations of Empirical Distance Correlations}\\

\noindent\textbf{\large Calculating the Eq. (\textcolor{red}{8})}

\begin{enumerate}
    \item We first derive the empirical distance correlation between the domain decoder output
$\hat f_{\theta_D}(x_i)$ and the feature encoder output
$\hat f_{\theta_E}(x_i)$ \emph{within} each subset of samples where $\mathbf{Y}=y$.
    \item We then aggregate those class-wise quantities using the empirical class probabilities so that the final estimate reflects the distribution of~$\mathbf{Y}$.
\end{enumerate}

Let
\[
\mathcal{Y}=\{y_1,\dots ,y_k\}
\quad\text{and}\quad
I_y=\bigl\{\,i\;:\;\mathbf{Y}^{\,i}=y \bigr\},\;
n_y=|I_y|,\;
n=\sum_{y\in\mathcal{Y}}n_y,\;
\hat p_{y}=n_y/n.
\]

Denote
\[
z_{D}^{i}=\hat f_{\theta_D}(x_{i}),
\qquad
z_{E}^{i}=\hat f_{\theta_E}(x_{i}),
\qquad
i=1,\dots ,n.
\]

or every $y\in\mathcal{Y}$ and every $i,j\in I_y$ set, define the Euclidean distance matrices
\begin{equation}
a_{i,j}^{(y)}
   \;=\;
   \bigl\|\,z_{D}^{i}-z_{D}^{j}\bigr\|_{2},
\qquad
b_{i,j}^{(y)}
   \;=\;
   \bigl\|\,z_{E}^{i}-z_{E}^{j}\bigr\|_{2}.
\end{equation}

\vspace{0.75em}
Calculating the row, column and grand means inside the class $y$:
\begin{align}
\overline{a}_{i\cdot}^{(y)}
   &=\frac{1}{n_y}\sum_{j\in I_y}a_{i,j}^{(y)}, &
\overline{b}_{i\cdot}^{(y)}
   &=\frac{1}{n_y}\sum_{j\in I_y}b_{i,j}^{(y)},\\
\overline{a}_{\cdot j}^{(y)}
   &=\frac{1}{n_y}\sum_{i\in I_y}a_{i,j}^{(y)}, &
\overline{b}_{\cdot j}^{(y)}
   &=\frac{1}{n_y}\sum_{i\in I_y}b_{i,j}^{(y)},\\
\overline{a}_{\cdot\cdot}^{(y)}
   &=\frac{1}{n_y^{2}}\sum_{i,j\in I_y}a_{i,j}^{(y)}, &
\overline{b}_{\cdot\cdot}^{(y)}
   &=\frac{1}{n_y^{2}}\sum_{i,j\in I_y}b_{i,j}^{(y)}.
\end{align}

We then calculate the doubly--centred distance matrices:

\[
A_{i,j}^{(y)}=a_{i,j}^{(y)}-\overline{a}_{i\cdot}^{(y)}
                        -\overline{a}_{\cdot j}^{(y)}
                        +\overline{a}_{\cdot\cdot}^{(y)},
\qquad
B_{i,j}^{(y)}=b_{i,j}^{(y)}-\overline{b}_{i\cdot}^{(y)}
                        -\overline{b}_{\cdot j}^{(y)}
                        +\overline{b}_{\cdot\cdot}^{(y)}.
\]

\vspace{0.3em}

The class–wise squared distance covariance and variances can be given as:

\begin{align}
\operatorname{dCov}^{2}_{n_{y}}
         \bigl(\hat f_{\theta_D},\hat f_{\theta_E}\mid \mathbf{Y}=y\bigr)
  &=\frac{1}{n_{y}^{2}}
    \sum_{i\in I_{y}}\sum_{j\in I_{y}}
    A_{i,j}^{(y)}\,B_{i,j}^{(y)}, \\[6pt]
\operatorname{dVar}^{2}_{n_{y}}
         \bigl(\hat f_{\theta_D}\mid \mathbf{Y}=y\bigr)
  &=\frac{1}{n_{y}^{2}}
    \sum_{i\in I_{y}}\sum_{j\in I_{y}}
    \bigl(A_{i,j}^{(y)}\bigr)^{2}, \\[6pt]
\operatorname{dVar}^{2}_{n_{y}}
         \bigl(\hat f_{\theta_E}\mid \mathbf{Y}=y\bigr)
  &=\frac{1}{n_{y}^{2}}
    \sum_{i\in I_{y}}\sum_{j\in I_{y}}
    \bigl(B_{i,j}^{(y)}\bigr)^{2}.
\end{align}

Because two independent samples fall \emph{jointly} into class $y$ with
probability~$\hat p_{y}^{\,2}$, the overall (conditional) squared
distance covariance is
\begin{align}
\operatorname{dCov}_{n}^{2}
   (\hat f_{\theta_D},\hat f_{\theta_E}\mid\mathbf{Y})
 &=\sum_{y\in\mathcal{Y}}
   \hat p_{y}^{\,2}\;
   \operatorname{dCov}^{2}_{n_y}
      (\hat f_{\theta_D},\hat f_{\theta_E}\mid\mathbf{Y}=y)\\
 &=\frac{1}{n^{2}}
   \sum_{y\in\mathcal{Y}}\;
   \sum_{i,j\in I_y}\!A_{i,j}^{(y)}B_{i,j}^{(y)}.
\end{align}

\noindent
Aggregating the variances with the \emph{same} weights gives
\begin{align}
\operatorname{dVar}_{n}^{2}
   (\hat f_{\theta_D}\mid\mathbf{Y})
 &=\sum_{y}\hat p_{y}^{\,2}\;
   \operatorname{dVar}^{2}_{n_y}
      (\hat f_{\theta_D}\mid\mathbf{Y}=y)
  =\frac{1}{n^{2}}
   \sum_{y}\;\sum_{i,j\in I_y}\!\bigl(A_{i,j}^{(y)}\bigr)^{2},\\
\operatorname{dVar}_{n}^{2}
   (\hat f_{\theta_E}\mid\mathbf{Y})
 &=\sum_{y}\hat p_{y}^{\,2}\;
   \operatorname{dVar}^{2}_{n_y}
      (\hat f_{\theta_E}\mid\mathbf{Y}=y)
  =\frac{1}{n^{2}}
   \sum_{y}\;\sum_{i,j\in I_y}\!\bigl(B_{i,j}^{(y)}\bigr)^{2}.
\end{align}

Finally, we have the empirical distance correlation conditioned on $\mathbf{Y}$

\begin{equation}
\operatorname{dCor}_{n}^{2}
   (\hat f_{\theta_D},\hat f_{\theta_E}\mid\mathbf{Y})
 \;=\;
 \frac{\operatorname{dCov}_{n}^{2}
         (\hat f_{\theta_D},\hat f_{\theta_E}\mid\mathbf{Y})}
      {\sqrt{\operatorname{dVar}_{n}^{2}
               (\hat f_{\theta_D}\mid\mathbf{Y})\;
             \operatorname{dVar}_{n}^{2}
               (\hat f_{\theta_E}\mid\mathbf{Y})}}.
\end{equation}

\vspace{0.75em}

\noindent
If the denominator is zero, we set
$\operatorname{dCor}_{n}^{2}(\hat f_{\theta_D},\hat f_{\theta_E}\mid\mathbf{Y})=0$. The statistic obeys
\(
0 \;\le\;
\operatorname{dCor}_{n}
   (\hat f_{\theta_D},\hat f_{\theta_E}\mid\mathbf{Y})
\;\le\; 1,
\)
and a value of~$0$ indicates no detectable dependence in the representations of samples once the outcome
$\mathbf{Y}$ is taken into account. \\

\noindent\textbf{\large Calculating the Eq. (\textcolor{red}{9})}\\

\begin{enumerate}
    \item We first compute the distance–covariance terms
          \emph{within} each subset of samples where
          $\mathbf{Y}=y$ \textbf{and} $\mathbf{D}_{S}=d_{S}$.
    \item We then aggregate those joint–class quantities using the
          empirical joint probabilities so that the final estimate
          reflects the observed distribution of $(\mathbf{Y},\mathbf{D}_{S})$.
\end{enumerate}

Let
\[
\mathcal{Y}=\{y_1,\dots ,y_k\},\qquad
\mathcal{D}_S=\{d_{S1},\dots ,d_{Sm}\},
\]
and define for every pair $(y,d_{S})$
\[
I_{y,d_{S}}=\bigl\{\,i\;:\;\mathbf{Y}^{\,i}=y,\; \mathbf{D}_{S}^{\,i}=d_{S}\bigr\},\;
n_{y,d_{S}}=|I_{y,d_{S}}|,\qquad
n=\sum_{y\in\mathcal{Y}}\sum_{d_{S}\in\mathcal{D}}n_{y,d_{S}},\qquad
\hat p_{y,d_{S}}=n_{y,d_{S}}/n.
\]

For $i=1,\dots ,n$ denote
\[
z_{G}^{i}=\hat f_{\theta_G}(x_{i}),
\qquad
z_{E}^{i}=\hat f_{\theta_E}(x_{i}).
\]

\vspace{0.8em}
For every $(y,d_{S})$ and all $i,j\in I_{y,d_{S}}$, define the pairwise distance matrices inside each joint class
\begin{equation}
a_{i,j}^{(y,d_{S})}=\|z_{G}^{i}-z_{G}^{j}\|_{2},
\qquad
b_{i,j}^{(y,d_{S})}=\|z_{E}^{i}-z_{E}^{j}\|_{2}.
\end{equation}

\vspace{0.8em}

We then calculate the row, column, and grand means (inside $(y,d_{S})$)

\begin{align}
\overline{a}_{i\cdot}^{(y,d_{S})}&=\tfrac{1}{n_{y,d_{S}}}\!\sum_{j\in I_{y,d_{S}}}a_{i,j}^{(y,d_{S})}, &
\overline{b}_{i\cdot}^{(y,d_{S})}&=\tfrac{1}{n_{y,d_{S}}}\!\sum_{j\in I_{y,d_{S}}}b_{i,j}^{(y,d_{S})},\\
\overline{a}_{\cdot j}^{(y,d_{S})}&=\tfrac{1}{n_{y,d_{S}}}\!\sum_{i\in I_{y,d_{S}}}a_{i,j}^{(y,d_{S})}, &
\overline{b}_{\cdot j}^{(y,d_{S})}&=\tfrac{1}{n_{y,d_{S}}}\!\sum_{i\in I_{y,d_{S}}}b_{i,j}^{(y,d_{S})},\\
\overline{a}_{\cdot\cdot}^{(y,d_{S})}&=\tfrac{1}{n_{y,d_{S}}^{2}}\!\sum_{i,j\in I_{y,d_{S}}}a_{i,j}^{(y,d_{S})}, &
\overline{b}_{\cdot\cdot}^{(y,d_{S})}&=\tfrac{1}{n_{y,d_{S}}^{2}}\!\sum_{i,j\in I_{y,d_{S}}}b_{i,j}^{(y,d_{S})}.
\end{align}

\vspace{0.8em}
The doubly–centred distance matrices are
\[
A_{i,j}^{(y,d_{S})}=a_{i,j}^{(y,d_{S})}
               -\overline{a}_{i\cdot}^{(y,d_{S})}
               -\overline{a}_{\cdot j}^{(y,d_{S})}
               +\overline{a}_{\cdot\cdot}^{(y,d_{S})},\quad
B_{i,j}^{(y,d_{S})}=b_{i,j}^{(y,d_{S})}
               -\overline{b}_{i\cdot}^{(y,d_{S})}
               -\overline{b}_{\cdot j}^{(y,d_{S})}
               +\overline{b}_{\cdot\cdot}^{(y,d_{S})}.
\]

\vspace{0.8em}
The joint–class squared distance covariance and variances (conditioned on $y,d_{S}$) are

\begin{align}
\operatorname{dCov}^{2}_{n_{y,d_{S}}}
   (\hat f_{\theta_G},\hat f_{\theta_E}\mid\mathbf{Y}=y,\mathbf{D}_{S}=d_{S})
 &=\frac{1}{n_{y,d_{S}}^{2}}
   \!\sum_{i,j\in I_{y,d_{S}}}\!A_{i,j}^{(y,d_{S})}B_{i,j}^{(y,d_{S})},\\[6pt]
\operatorname{dVar}^{2}_{n_{y,d_{S}}}
   (\hat f_{\theta_G}\mid\mathbf{Y}=y,\mathbf{D}_{S}=d_{S})
 &=\frac{1}{n_{y,d_{S}}^{2}}
   \!\sum_{i,j\in I_{y,d_{S}}}\!\bigl(A_{i,j}^{(y,d_{S})}\bigr)^{2},\\[6pt]
\operatorname{dVar}^{2}_{n_{y,d_{S}}}
   (\hat f_{\theta_E}\mid\mathbf{Y}=y,\mathbf{D}_{S}=d_{S})
 &=\frac{1}{n_{y,d_{S}}^{2}}
   \!\sum_{i,j\in I_{y,d_{S}}}\!\bigl(B_{i,j}^{(y,d_{S})}\bigr)^{2}.
\end{align}

\vspace{0.8em}

Because two independent samples land jointly in $(y,d_{S})$ with probability
$\hat p_{y,d_{S}}^{\,2}$, the conditional squared distance covariance is
\begin{align}
\operatorname{dCov}_{n}^{2}
  (\hat f_{\theta_G},\hat f_{\theta_E}\mid\mathbf{Y},\mathbf{D}_{S})
 &=\sum_{y\in\mathcal{Y}}\sum_{d_{S}\in\mathcal{D}}
   \hat p_{y,d_{S}}^{\,2}\;
   \operatorname{dCov}^{2}_{n_{y,d_{S}}}
     (\hat f_{\theta_G},\hat f_{\theta_E}\mid\mathbf{Y}=y,\mathbf{D}_{S}=d_{S})\\
 &=\frac{1}{n^{2}}
   \sum_{y,d_{S}}\;\sum_{i,j\in I_{y,d_{S}}}
   A_{i,j}^{(y,d_{S})}\,B_{i,j}^{(y,d_{S})}.
\end{align}

Aggregating the variances with the same weights yields
\begin{align}
\operatorname{dVar}_{n}^{2}
   (\hat f_{\theta_G}\mid\mathbf{Y},\mathbf{D}_{S})
 &=\sum_{y,d_{S}}\hat p_{y,d_{S}}^{\,2}\;
   \operatorname{dVar}^{2}_{n_{y,d_{S}}}
     (\hat f_{\theta_G}\mid\mathbf{Y}=y,\mathbf{D}_{S}=d_{S})
  =\frac{1}{n^{2}}
   \sum_{y,d_{S}}\sum_{i,j\in I_{y,d_{S}}}
     \bigl(A_{i,j}^{(y,d_{S})}\bigr)^{2},\\
\operatorname{dVar}_{n}^{2}
   (\hat f_{\theta_E}\mid\mathbf{Y},\mathbf{D}_{S})
 &=\sum_{y,d_{S}}\hat p_{y,d_{S}}^{\,2}\;
   \operatorname{dVar}^{2}_{n_{y,d_{S}}}
     (\hat f_{\theta_E}\mid\mathbf{Y}=y,\mathbf{D}_{S}=d_{S})
  =\frac{1}{n^{2}}
   \sum_{y,d_{S}}\sum_{i,j\in I_{y,d_{S}}}
     \bigl(B_{i,j}^{(y,d_{S})}\bigr)^{2}.
\end{align}
Finally, we get the empirical distance correlation conditioned on {$(\mathbf{Y},\mathbf{D}_{S})$}

\begin{equation}
\operatorname{dCor}_{n}^{2}
   (\hat f_{\theta_G},\hat f_{\theta_E}\mid\mathbf{Y},\mathbf{D}_{S})
 =\frac{\operatorname{dCov}_{n}^{2}
         (\hat f_{\theta_G},\hat f_{\theta_E}\mid\mathbf{Y},\mathbf{D}_{S})}
        {\sqrt{\operatorname{dVar}_{n}^{2}
                 (\hat f_{\theta_G}\mid\mathbf{Y},\mathbf{D}_{S})\;
               \operatorname{dVar}_{n}^{2}
                 (\hat f_{\theta_E}\mid\mathbf{Y},\mathbf{D}_{S})}}.
\end{equation}

\noindent
If the denominator is zero we set
$\operatorname{dCor}_{n}^{2}(\hat f_{\theta_G},\hat f_{\theta_E}\mid\mathbf{Y},\mathbf{D}_{S})=0$.
By construction
\(
0\;\le\;
\operatorname{dCor}_{n}
   (\hat f_{\theta_G},\hat f_{\theta_E}\mid\mathbf{Y},\mathbf{D}_{S})
\;\le\; 1,
\)
and a value of $0$ implies no detectable dependence between the representations of samples after conditioning on both
$\mathbf{Y}$ and $\mathbf{D}_{S}$. \\

\noindent\textbf{\Large  \textcolor{red}{F}. Datasets and Implementation Details}\\

Dataset construction scripts and implementation details can be found at \url{https://github.com/Supltz/FairDG}. 

\begin{itemize}
    \item \textbf{\large Datasets}
\end{itemize}

\begin{itemize}
    \item \textbf{CelebA}
\end{itemize}

 \textbf{CelebA} is a widely used benchmark for fairness in facial attribute classification. While it is originally a multi-label classification dataset with 40 binary facial attributes, the FairDG problem we investigate in this paper focuses on multi-class classification with a multi-group sensitive attribute and requires splitting into multiple domains (at least four) for training, validation, and testing. To simulate the FairDG setting, we carefully select and reconstruct the labels. The classification task is defined as predicting hair color from {\textit{black hair}, \textit{brown hair}, \textit{blond hair}}, ensuring the classes are mutually exclusive (each face image belongs to only one hair color). Domain variables are defined by hairstyle types: {\textit{wavy hair}, \textit{straight hair}, \textit{bangs}, \textit{receding hairlines}}, which are also mutually exclusive. Here, \textit{bangs} is designated as the unseen target domain for testing, \textit{receding hairlines} is used for validation, and the remaining domains are used for training. The sensitive attribute is defined as the intersection of perceived gender and age group, resulting in four mutually exclusive groups: {\textit{male-young}, \textit{female-young}, \textit{male-old}, \textit{female-old}}. This construction yields a total of 65,372 face images, with 53,845 for training, 3,810 for validation, and 7,717 for testing.

 \begin{itemize}
    \item \textbf{AffectNet}
\end{itemize}

\textbf{AffectNet} is the largest in-the-wild facial expression dataset, containing 286,339 face images annotated with seven categories: {\textit{Happiness}, \textit{Sadness}, \textit{Neutral}, \textit{Fear}, \textit{Anger}, \textit{Surprise}, \textit{Disgust}}. While the original dataset provides only facial expression annotations, we incorporate perceived age and race annotations from Hu et al. \cite{hu2025rethinking}. The domain variable is \textit{age}, which we regroup into five categories: {0–9, 10–29, 30–49, 50–69, 70+}. The sensitive attribute is \textit{race}, with four groups: {\textit{White}, \textit{Black}, \textit{East Asian}, \textit{Indian}}. In our setup, the 0–9 age group serves as the unseen target domain for testing, the 10–29 group is used for validation, and the remaining groups are used for training. This construction yields 120,257 images for training, 132,439 for validation, and 33,634 for testing.

 \begin{itemize}
    \item \textbf{Jigsaw}
\end{itemize}
 
The \textbf{Jigsaw} dataset focuses on toxicity classification in text. The task involves predicting toxicity levels for comments labeled as {\textit{non-toxic}, \textit{toxic}, \textit{severe toxic}}. The original dataset provides toxicity intensity scores, which we discretize as follows: a score of 0 is labeled \textit{non-toxic}, scores in the range (0, 0.1] are labeled \textit{toxic}, and scores greater than 0.1 are labeled \textit{severe toxic}. Toxicity types in the original dataset are multi-label; we filter the samples to ensure these categories are mutually exclusive. These toxicity types serve as domain variables: {\textit{Obscene}, \textit{Identity attack}, \textit{Insult}, \textit{Threat}}. Here, \textit{Identity attack} is used as the unseen target domain for testing, \textit{Threat} is used for validation, and the remaining domains are used for training. The sensitive attribute is defined as the presence of gender-related terms: {\textit{male}, \textit{female}, \textit{transgender}}. Since these attributes are multi-label in the original dataset, we filter them to make the groups mutually exclusive. After this filtering and reconstruction, the dataset contains 16,188 comments, split into 10,020 for training, 1,113 for validation, and 5,055 for testing. \\

\begin{itemize}
    \item \textbf{\large Implementation Details}
\end{itemize}

\begin{table*}[]
\centering
\caption{Summary of training hyperparameters, including batch size, learning rate, momentum, weight decay, and early stopping patience for each dataset.}

  \label{tab:freq}
  \setlength{\tabcolsep}{3pt} 
  \renewcommand{\arraystretch}{0.7} 
\begin{tabular}{>{\raggedleft\arraybackslash}c|c|c|c|c|c}
  \toprule
  Datasets & Batch Size & Learning Rate & Momentum & Weight Decay & Patience\\
  \midrule
  \hfill CelebA & 32 & 2e-4 & 0.8 & 5e-3 & 5\\
  \midrule
  \hfill AffectNet & 64 & 1e-4 & 0.9 & 5e-5 & 5\\
  \midrule
  \hfill Jigsaw & 8 & 1e-4 & 0.9 & 5e-4 & 5\\
  \midrule
\end{tabular}
\end{table*}

The trade-off parameter $\lambda$ was varied in the range [0, 1) with a step size of 0.01 ($N = 100$). 
We conduct a toy experiment on the CelebA dataset with $N = 10, 20, 25, 40, 50, 80, 100$ (uniform distribution). As shown in the Table \textcolor{red}{8}, we observe that the HVI is already quite stable at $ N = 80$. 100 is a safer choice across different experimental settings. \\

\begin{table}[h!]
    \centering
    \caption{HVI values across different sample sizes $N$ in the CelebA toy experiment.}
    \begin{tabular}{c|ccccccc}
        \toprule
        $N$ & 10 & 20 & 25 & 40 & 50 & 80 & 100 \\
        \midrule
        HVI & 67.2 & 69.8 & 71.4 & 73.7 & 75.1 & 75.4 & 75.4 \\
        \bottomrule
    \end{tabular}
\end{table}

We also demonstrate the role of $\lambda$ as a trade-off coefficient using the same toy experiment. As shown in Fig. \textcolor{red}{4}, increasing $\lambda$ generally reduces EOD violations (i.e., improves fairness), but also leads to lower accuracy, revealing the expected trade-off. However, the trend is not strictly monotonic as highlighted by the non-monotonic points in the figure. We attribute this to stochastic errors in both ERM and the dCor estimate, which may not fully capture expected risk and statistical independence due to limited sample size. Nonetheless, fewer than 10\% of the $\lambda$ values deviate from the trend. We expect this percentage to drop with more data and leave the deeper investigation to future work.

\begin{figure}[ht]
    \centering
    \begin{subfigure}[b]{0.48\textwidth}
        \centering
        \includegraphics[width=\textwidth]{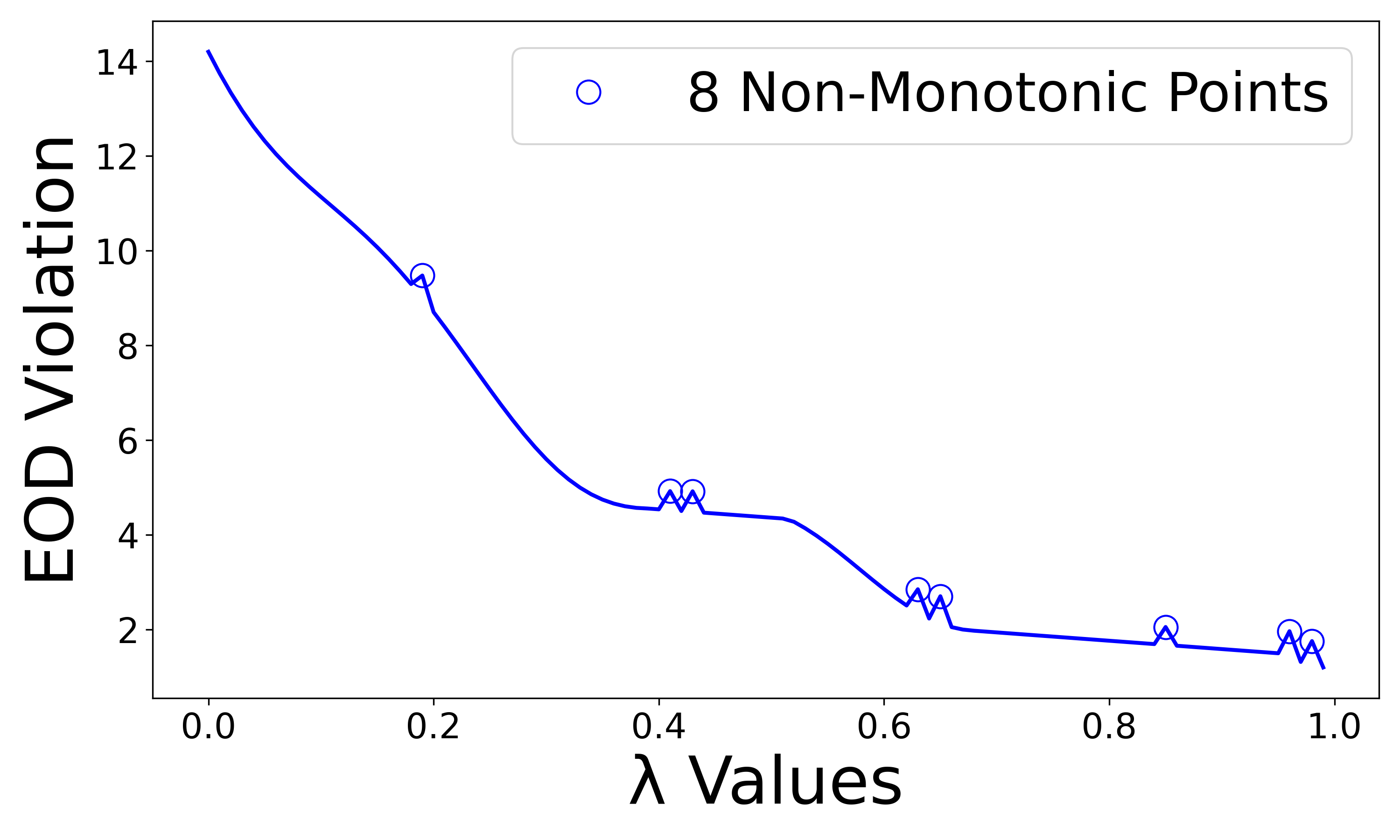}
        \label{fig:eod}
    \end{subfigure}
    \hfill
    \begin{subfigure}[b]{0.48\textwidth}
        \centering
        \includegraphics[width=\textwidth]{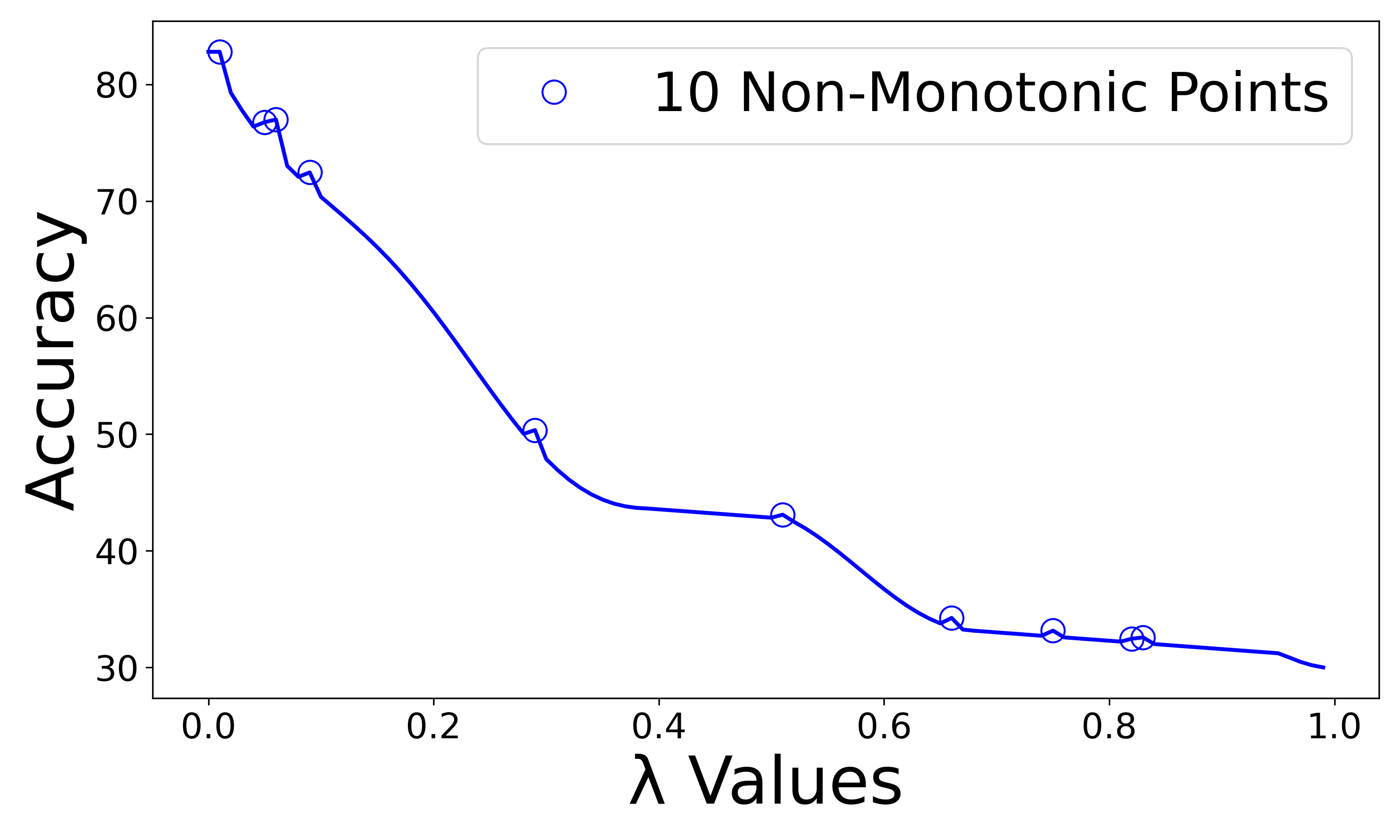}
        \label{fig:accuracy}
    \end{subfigure}
    \caption{EOD Violation and Accuracy with respect to the trade-off coefficient $\lambda$ (CelebA dataset).}
    \label{fig:comparison}
\end{figure}

\newpage

\noindent\textbf{\Large  \textcolor{red}{G}. Table \textcolor{red}{4} with Variances and the Corresponding HVI Values of the Pareto Fronts in Figure \textcolor{red}{3}}\\

\begin{table}[h]
\centering
\caption{\textbf{Corresponding to the Table \textcolor{red}{4} in the main paper with variances included.} Comparison of existing methods for \((V_{\text{opt}}, U_{\text{opt}})\) evaluation. DG methods are marked with \textcolor{red}{$\dagger$}, fairness-only methods with \textcolor{red}{$*$}, and FairDG methods with \textcolor{red}{$\dagger*$}. Higher Acc (\%) reflects better utility, while lower EOD (\%) and EO (\%) indicate better fairness. The best-performing method is shown in bold and underlined; the second-best is underlined.}
\setlength{\tabcolsep}{6pt}
\renewcommand{\arraystretch}{1.2}
{\fontsize{9pt}{9pt}\selectfont
\begin{tabular}{c|c|c|c|c|c|c|c|c|c}
\hline
  Dataset &
  \multicolumn{3}{c|}{CelebA} &
  \multicolumn{3}{c|}{AffectNet} &
  \multicolumn{3}{c}{Jigsaw} \\ \hline
  Methods &
  \multicolumn{1}{c|}{Acc \textcolor{red}{$\uparrow$}} &
  \multicolumn{1}{c|}{EOD \textcolor{blue}{\textdownarrow}} &
  \multicolumn{1}{c|}{EO \textcolor{blue}{\textdownarrow}} &
  \multicolumn{1}{c|}{Acc \textcolor{red}{$\uparrow$}} &
  \multicolumn{1}{c|}{EOD \textcolor{blue}{\textdownarrow}} &
  \multicolumn{1}{c|}{EO \textcolor{blue}{\textdownarrow}} &
  \multicolumn{1}{c|}{Acc \textcolor{red}{$\uparrow$}} &
  \multicolumn{1}{c|}{EOD \textcolor{blue}{\textdownarrow}} &
  \multicolumn{1}{c}{EO \textcolor{blue}{\textdownarrow}} \\ \hline
\rowcolor{gray!20} 
ERM (Ours) & 82.8 ±0.8 & 14.2 ±0.6 & 10.5 ±1.1 & 62.8 ±1.2 & 15.0 ±0.9 & 10.1 ±1.3 & 82.4 ±0.5 & 17.8 ±1.2 & 11.1 ±0.7 \\
\rowcolor{gray!20} 
\textcolor{red}{$\dagger$} ERM+SDI (Ours) & \textbf{\underline{89.6}} ±0.4 & 13.5 ±1.2 & 11.5 ±1.0 & \textbf{\underline{69.8}} ±1.1 & 15.2 ±0.8 & 11.1 ±0.5 & \textbf{\underline{87.4}} ±0.7 & 18.8 ±0.9 & 11.6 ±0.6 \\
\textcolor{red}{$\dagger$} DANN & 88.6 ±1.0 & 15.5 ±0.7 & 14.4 ±1.2 & 66.8 ±0.5 & 16.5 ±0.9 & 12.4 ±0.8 & 84.2 ±0.9 & 17.7 ±1.3 & 12.1 ±0.4 \\
\textcolor{red}{$\dagger$} CORAL & 87.6 ±0.5 & 14.6 ±1.4 & 14.5 ±1.1 & 67.3 ±1.2 & 15.6 ±0.6 & 14.5 ±0.7 & 83.1 ±0.8 & 18.6 ±1.0 & 13.2 ±0.9 \\
\textcolor{red}{$\dagger$} MMD-AAE & 88.4 ±1.3 & 16.5 ±0.5 & 13.2 ±0.7 & 68.4 ±0.6 & 17.5 ±0.8 & 15.2 ±1.1 & 84.4 ±0.7 & 16.8 ±1.4 & 10.7 ±0.5 \\
\textcolor{red}{$\dagger$} DDG & 86.9 ±0.9 & 17.2 ±1.0 & 14.2 ±0.6 & \underline{69.4} ±0.8 & 18.2 ±0.5 & 14.2 ±1.3 & 84.7 ±1.1 & 18.5 ±0.4 & 11.9 ±0.7 \\ \hline
\rowcolor{gray!20} 
\textcolor{red}{$*$} ERM+Fair (Ours) & 79.4 ±1.1 & 13.8 ±0.5 & 10.4 ±0.9 & 59.4 ±0.7 & 14.8 ±1.3 & 8.4 ±0.6 & 81.4 ±0.4 & 15.5 ±0.8 & 10.9 ±1.2 \\
\textcolor{red}{$*$} LNL & 72.4 ±1.0 & 13.2 ±1.1 & 10.4 ±0.5 & 61.4 ±0.9 & 13.6 ±0.8 & 9.6 ±1.2 & 77.4 ±0.7 & 17.5 ±0.6 & 7.1 ±1.1 \\
\textcolor{red}{$*$} MaxEnt-ARL & 71.5 ±0.8 & 13.6 ±1.2 & 9.0 ±0.4 & 61.5 ±1.3 & 13.8 ±1.0 & 10.0 ±0.9 & 78.7 ±0.6 & 16.5 ±0.7 & 7.7 ±1.2 \\
\textcolor{red}{$*$} FairHSIC & 82.4 ±0.9 & 12.5 ±0.6 & 9.3 ±0.8 & 62.4 ±0.5 & 12.6 ±1.4 & 9.8 ±1.1 & 79.4 ±0.8 & 15.6 ±1.0 & 9.8 ±0.6 \\
\textcolor{red}{$*$} U-FaTE & 69.5 ±0.7 & 14.1 ±1.3 & 8.2 ±0.9 & 59.5 ±0.6 & 14.6 ±0.8 & 6.4 ±1.1 & 79.7 ±0.5 & 16.1 ±1.2 & 8.7 ±0.4 \\ \hline
\textcolor{red}{$\dagger*$} FEDORA & 85.7 ±1.2 & 10.6 ±0.6 & 8.1 ±0.4 & 65.8 ±0.9 & 10.2 ±1.1 & \underline{5.9} ±0.5 & 84.4 ±0.7 & 12.6 ±1.3 & 6.1 ±0.8 \\
\textcolor{red}{$\dagger*$} FATDM-StarGAN & 85.4 ±0.5 & 10.9 ±1.0 & 6.7 ±1.2 & 65.6 ±0.7 & \underline{9.8} ±0.8 & 6.3 ±0.4 & 85.7 ±0.9 & \underline{12.3} ±1.1 & 5.7 ±0.5 \\
\rowcolor{gray!20} 
\textcolor{red}{$\dagger*$} Ours-S & 84.3 ±0.8 & 11.8 ±0.7 & \underline{5.8} ±0.6 & 64.5 ±1.2 & 10.8 ±0.9 & 6.3 ±1.1 & 84.6 ±0.5 & 12.4 ±0.8 & \underline{4.9} ±1.0 \\
\rowcolor{gray!20} 
\textbf{\textcolor{red}{$\dagger*$} Ours} & \underline{88.7} ±0.6 & \textbf{\underline{8.2}} ±1.1 & \textbf{\underline{3.1}} ±0.5 & 68.9 ±1.3 & \textbf{\underline{8.4}} ±1.3 & \textbf{\underline{4.3}} ±0.8 & \underline{86.3} ±1.1 & \textbf{\underline{9.7}} ±0.7 & \textbf{\underline{3.7}} ±0.9 \\ \hline
\end{tabular}
}
\end{table}

\begin{table}[h]
\centering
\caption{\textbf{Corresponding to the Figure \textcolor{red}{3} in the main paper with variances included.} Comparison of existing methods for $\mathcal{P}_{\text{norm}}$ evaluation. Fairness methods are marked with \textcolor{red}{$*$}, and FairDG methods with \textcolor{red}{$\dagger*$}. Higher HVI (\%) indicates a better Pareto front. The best-performing method is shown in bold and underlined; the second-best is underlined.}
\setlength{\tabcolsep}{7pt}
\renewcommand{\arraystretch}{1.4}
{\fontsize{10pt}{9pt}\selectfont
\begin{tabular}{c|cc|cc|cc}
\hline
  Dataset &
  \multicolumn{2}{c|}{CelebA} &
  \multicolumn{2}{c|}{AffectNet} &
  \multicolumn{2}{c}{Jigsaw} \\ \hline
  Methods &
  \multicolumn{1}{c|}{HVI (EOD) \textcolor{red}{$\uparrow$}} & 
  \multicolumn{1}{l|}{HVI (EO) \textcolor{red}{$\uparrow$}} &
  \multicolumn{1}{l|}{HVI (EOD) \textcolor{red}{$\uparrow$}} &
  \multicolumn{1}{l|}{HVI (EO) \textcolor{red}{$\uparrow$}} &
  \multicolumn{1}{l|}{HVI (EOD) \textcolor{red}{$\uparrow$}} &
  \multicolumn{1}{l}{HVI (EO) \textcolor{red}{$\uparrow$}} \\ \hline
  \rowcolor{gray!20} 
\textcolor{red}{$*$} ERM+Fair (Ours) & 56.9 ±0.8 & 51.1 ±1.2 & 52.8 ±0.9 & 57.8 ±1.0 & 59.4 ±1.1 & 53.9 ±0.5 \\
\textcolor{red}{$*$} LNL & 54.8 ±0.6 & 50.5 ±1.3 & 58.3 ±0.7 & 56.8 ±0.4 & 50.2 ±1.0 & 59.3 ±1.2 \\
\textcolor{red}{$*$} MaxEnt-ARL & 54.2 ±1.0 & 58.7 ±0.5 & 58.5 ±1.1 & 49.9 ±0.8 & 53.5 ±1.3 & 58.2 ±0.9 \\
\textcolor{red}{$*$} FairHSIC & 60.4 ±0.9 & 56.8 ±0.7 & 60.1 ±1.2 & 54.8 ±0.6 & 54.7 ±1.1 & 53.9 ±1.0 \\
\textcolor{red}{$*$} U-FaTE & 51.2 ±0.4 & 58.6 ±1.0 & 53.0 ±1.3 & 58.2 ±0.9 & 54.2 ±0.5 & 55.7 ±1.1 \\ \hline
\textcolor{red}{$\dagger*$} FEDORA & 70.4 ±0.5 & 69.5 ±0.8 & \underline{72.8} ±1.1 & \underline{71.5} ±0.7 & 68.8 ±1.0 & 69.6 ±0.9 \\
\textcolor{red}{$\dagger*$} FATDM-StarGAN & 70.0 ±1.2 & 71.8 ±0.4 & \underline{72.8} ±0.5 & 68.0 ±1.0 & \underline{71.2} ±0.8 & 71.3 ±1.3 \\
\rowcolor{gray!20} 
\textcolor{red}{$\dagger*$} Ours-S & 68.1 ±0.6 & \underline{72.6} ±1.1 & 69.3 ±0.9 & 68.6 ±0.7 & 69.3 ±0.4 & \underline{72.2} ±1.2 \\
\rowcolor{gray!20} 
\textbf{\textcolor{red}{$\dagger*$} Ours} & \textbf{\underline{75.4}} ±0.5 & \textbf{\underline{78.3}} ±0.8 & \textbf{\underline{76.4}} ±0.6 & \textbf{\underline{74.9}} ±1.0 & \textbf{\underline{75.8}} ±0.7 & \textbf{\underline{75.7}} ±1.1 \\ \hline
\end{tabular}
}
\end{table}

\end{document}